\documentclass{article}

\usepackage[utf8]{inputenc}

\RequirePackage[OT1]{fontenc}
\RequirePackage{amsthm,amsmath,amsfonts,amssymb,dsfont}
\RequirePackage[numbers]{natbib}
\RequirePackage[colorlinks,citecolor=blue,urlcolor=blue]{hyperref}

\usepackage{graphicx}

\usepackage{xcolor}
\usepackage{breakcites} % repare pb de mise à la ligne des references
\newtheorem{theorem}{Theorem}
\newtheorem{assumption}[theorem]{Assumption}

\newtheorem{definition}[theorem]{Definition}

\newtheorem{lemma}[theorem]{Lemma}

\newtheorem{proposition}[theorem]{Proposition}

\newcommand{\cA}{\mathcal{A}}

\newcommand{\cC}{\mathcal{C}}

\newcommand{\cE}{\mathcal{E}}
\newcommand{\cF}{\mathcal{F}}

\newcommand{\cI}{\mathcal{I}}

\newcommand{\cP}{\mathcal{P}}

\newcommand{\cR}{\mathcal{R}}
\newcommand{\cS}{\mathcal{S}}
\newcommand{\cT}{\mathcal{T}}
\newcommand{\cU}{\mathcal{U}}
\newcommand{\cV}{\mathcal{V}}

\newcommand{\bE}{\mathbb{E}}
\newcommand{\bP}{\mathbb{P}}
\newcommand{\N}{\mathbb{N}}
\newcommand{\R}{\mathbb{R}}

\def \ind{\mathds{1}}

\newcommand{\Jc}{\mathrm{J}_{\cC}}  
\newcommand{\KL}{\mathrm{KL}}
\newcommand{\Kl}{\mathrm{K}}
\newcommand{\Kd}{\mathrm{d}}
\newcommand{\JJ}{\mathrm{J}}

\newcommand{\thetaa}{\bm{\theta}_{-a}}
\newcommand{\thetaB}{\bm{\theta}}

\usepackage{bm}

\begin{document}

%\begin{frontmatter}
\title{On Bayesian index policies for sequential resource allocation}%\thanksref{T1}}
%\runtitle{On Bayesian index policies}
%\thankstext{T1}{Footnote to the title with the ``thankstext'' command.}

%\begin{aug}
\author{Emilie Kaufmann}
%\thanksref{t1}\author{\fnms{Second} \snm{Author}\thanksref{t3,m1,m2}\ead[label=e2]{second@somewhere.com}}
%\and
%\author{\fnms{Third} \snm{Author}\thanksref{t1,m2}
%\ead[label=e3]{third@somewhere.com}
%\ead[label=u1,url]{http://www.foo.com}}

%\thankstext{t1}{ALICIA, Spadro, Gap? }
%\thankstext{t2}{First supporter of the project}
%\thankstext{t3}{Second supporter of the project}
%\runauthor{E. Kaufmann}

\date{CNRS \& Univ. Lille, UMR 9189 (CRIStAL), Inria SequeL \\ \texttt{emilie.kaufmann@univ-lille1.fr}}

\maketitle

\begin{abstract}

This paper is about index policies for minimizing (frequentist) regret in a stochastic multi-armed bandit model, inspired by a Bayesian view on the problem. Our main contribution is to prove that the Bayes-UCB algorithm, which relies on quantiles of posterior distributions, is asymptotically optimal when the reward distributions belong to a one-dimensional exponential family, for a large class of prior distributions. We also show that the Bayesian literature gives new insight on what kind of exploration rates could be used in frequentist, UCB-type algorithms. Indeed, approximations of the Bayesian optimal solution or the Finite Horizon Gittins indices provide a justification for the kl-UCB$^+$ and kl-UCB-H$^+$ algorithms, whose asymptotic optimality is also established.  

\end{abstract}

\section{Introduction}

This paper presents new analyses of Bayesian flavored strategies for sequential resource allocation in an unknown, stochastic environment modeled as a multi-armed bandit. A \emph{stochastic multi-armed bandit model} %(or bandit model) 
is a set of $K$ probability distributions, $\cV_1,\dots,\cV_K$, called arms, with which an agent interacts in a sequential way. At round $t$, the agent, who does not know the arms' distributions, chooses an arm $A_t$. The draw of this arm produces an independent sample $X_t$ from the associated probability distribution $\cV_{A_t}$, often interpreted as a reward. Indeed, the arms can be viewed as those of different slot machines, also called \emph{one-armed bandits}, generating rewards according to some underlying probability distribution.

In several applications that range from the motivating example of clinical trials \cite{Thompson33} to the more modern motivation of online advertisement (e.g., \cite{LiChapelle11}), the goal of the agent is to adjust his strategy $\cA=(A_t)_{t\in\N}$, also called a \emph{bandit algorithm}, in order to maximize the rewards accumulated during his interaction with the bandit model. The adopted strategy has to be sequential, in the sense that the next arm to play is chosen based on past observations: letting $\cF_t= \sigma(A_1,X_1,\dots,A_t,X_t)$ be the $\sigma$-field generated by the observations up to round $t$, $A_t$ is $\sigma(\cF_{t-1},U_t)$-measurable, where $U_t$ is a uniform random variable independent from $\cF_{t-1}$ (as algorithms may be randomized). 

More precisely, the goal is to design a sequential strategy maximizing the expectation of the sum of rewards up to some horizon $T$. If $\mu_1,\dots,\mu_K$ denote the means of the arms, and $\mu^*=\max_a \mu_a$, this is equivalent to minimizing the \emph{regret}, defined as the expected difference between the reward accumulated by an oracle strategy always playing the best arm, and the reward accumulated by a strategy $\cA$: 
\begin{equation}\mathrm{R}(T,\cA) :=\bE\left[T \mu^* - \sum_{t=1}^T X_t\right] = \bE\left[\sum_{t=1}^T (\mu^* - \mu_{A_t})\right].\label{def:RegretGene}\end{equation}
The expectation is taken with respect to the randomness in the sequence of successive rewards from each arm $a$, denoted by $(Y_{a,s})_{s\in\N}$, and the possible randomization of the algorithm, $(U_t)_{t}$. We denote by $N_a(t) = \sum_{s=1}^t \ind_{(A_s=a)}$ the number of draws from arm $a$ at the end of round $t$, so that $X_t = Y_{A_t,N_{A_t}(t)}$.

This paper focuses on good strategies in \emph{parametric} bandit models, in which the distribution of arm $a$ depends on some parameter $\theta_a$: we write $\cV_a=\nu_{\theta_a}$. Like in every parametric model, two different views can be adopted. In the frequentist view, $\bm{\theta}=(\theta_1,\dots,\theta_K)$ is an unknown parameter. In the Bayesian view, $\bm{\theta}$ is a random variable, drawn from a prior distribution $\Pi$. More precisely, we define $\bP_{\bm \theta}$ (resp. $\bE_{\bm \theta}$) the probability (resp. expectation) under the probabilistic model in which for all $a$, $(Y_{a,s})_{s\in \N}$ is i.i.d. distributed under $\nu_{\theta_a}$ and $\bP^{\Pi}$ (resp. $\bE^\Pi$) the probability (resp. expectation) under the probabilistic model in which for all $a$ $(Y_{a,s})_{s \in \N}$ is i.i.d. conditionally to $\theta_a$ with conditional distribution $\nu_{\theta_a}$, and $\bm \theta \sim \Pi$.
The expectation in \eqref{def:RegretGene} can thus be taken under either of these two probabilistic models. In the first case this leads to the notion of frequentist regret, which depends on $\bm \theta$: 
\begin{equation}\mathrm{R}_{\bm \theta}(T,\cA) : = \bE_{\bm \theta}\left[\sum_{t=1}^T (\mu^* - \mu_{A_t})\right] = \sum_{a=1}^{K} (\mu^*-\mu_a) \bE_{\bm \theta}[N_a(T)].\label{def:RegretFreq}\end{equation}
In the second case, this leads to the notion of Bayesian regret, sometimes called \emph{Bayes risk} in the literature (see \cite{Lai87}), which depends on the prior distribution $\Pi$:
\begin{equation}\mathrm{\cR}_{\Pi}(T,\cA) : = \bE^{\Pi}\left[\sum_{t=1}^T (\mu^* - \mu_{A_t})\right] = \int \mathrm{R}_{\bm \theta}(T,\cA)d\Pi(\bm{\theta}).\label{def:RegretBay}\end{equation}

The first bandit strategy was introduced by Thompson in 1933 \cite{Thompson33} in a Bayesian framework, and a large part of the early work on bandit models is adopting the same perspective \cite{Bradtal56,BellmanBay56,Gittins79,BerryFristedt85}. Indeed, as Bayes risk minimization has an \emph{exact}---yet often intractable---solution, finding ways to efficiently compute this solution has been an important line of research. 
Since 1985 and the seminal work of Lai and Robbins \cite{LaiRobbins85bandits}, there is also a precise characterization of good bandit algorithms in a frequentist sense. They show that for any \emph{uniformly efficient policy} $\cA$ (i.e. such that for all $\bm \theta$, $\mathrm{R}_{\bm \theta}(T,\cA) = o(T^\alpha)$ for all $\alpha \in ]0,1]$), the number of draws of any sub-optimal arm $a$ ($\mu_a < \mu^*$) is asymptotically lower bounded as follows:
\begin{equation}\liminf_{T \rightarrow \infty} \frac{\bE_{\bm \theta}[N_a(T)]}{\log T} \geq \frac{1}{\mathrm{KL}(\nu_{\theta_a},\nu_{\theta^*})},\label{equ:LR}\end{equation}
where $\text{KL}(\nu,\nu')$ denotes the Kullback-Leibler divergence between the distributions $\nu$ and $\nu'$. From \eqref{def:RegretFreq}, this yields a lower bound on the regret.

This result holds for simple parametric bandit models, including exponential family bandit models presented in Section~\ref{sec:Exponential}, that will be our main focus in this paper. It paved the way to a new line of research, aimed at building \emph{asymptotically optimal} strategies, that is, strategies matching the lower bound \eqref{equ:LR} for some classes of distributions. Most of the algorithms proposed since then belong to the family of \emph{index policies}, that compute at each round one index per arm, depending on the history of rewards observed from this arm only, and select the arm with largest index. More precisely, they are UCB-type algorithms, building confidence intervals for the means of the arms and choosing as an index for each arm the associated Upper Confidence Bound (UCB). The design of the confidence intervals has been successively improved \cite{Lai87,Agrawal95,Aueral02,Audibertal09UCBV,Audibertal10MOSS,HondaTakemura10, KLUCBJournal} so as to obtain simple index policies for which non-asymptotic upper bound on the regret can be given. Among them, the kl-UCB algorithm \cite{KLUCBJournal} matches the lower bound \eqref{equ:LR} for exponential family bandit models. As they use confidence intervals on unknown parameters, all these index policies are based on \emph{frequentist tools}. Nevertheless, it is interesting to note that the first index policy was introduced by Gittins in 1979 \cite{Gittins79} to solve a Bayesian multi-armed bandit problem and is based on \emph{Bayesian tools}, i.e. on exploiting the posterior distribution on the parameter of each arm.    

However, tools and objectives can be separated: one can compute the Bayes risk of an algorithm based on frequentist tools, or the (frequentist) regret of an algorithm based on Bayesian tools. In this paper, we focus on the latter and advocate the use of index policies inspired by Bayesian tools for minimizing regret, in particular the Bayes-UCB algorithm \cite{AISTATS12}, which is based on quantiles of the posterior distributions on the means. Our main contribution is to prove that this algorithm is asymptotically optimal, i.e. that it matches the lower bound \eqref{equ:LR}, for any exponential bandit model and for a large class of prior distributions. Our analysis relies on two new ingredients: tight bounds on the tail of posterior distributions (Lemma~\ref{lem:DeterministicULB}), and a self-normalized deviation inequality featuring an exploration rate that decreases with the number of observations (Lemma~\ref{lem:TermA}). This last tool also allows us to prove the asymptotic optimality of two variants of kl-UCB, called kl-UCB$^+$ and kl-UCB-H$^+$, that display improved empirical performance. Interestingly, the alternative exploration rate used by these two algorithms is already suggested by asymptotic approximations of the Bayesian exact solution or the Finite-Horizon Gittins indices. 

The paper is structured as follows. Section~\ref{sec:Exponential} introduces the class of exponential family bandit models that we consider in the rest of the paper, and the associated frequentist and Bayesian tools. In Section~\ref{sec:BayesUCB}, we present the Bayes-UCB algorithm, and give a proof of its asymptotic optimality. We introduce kl-UCB$^+$ and kl-UCB-H$^+$ in Section~\ref{sec:KLUCBPlus}, in which we prove their asymptotic optimality and also exhibit connections with existing Bayesian policies. In Section~\ref{sec:Experiments}, we illustrate numerically the good performance of our three asymptotically optimal, Bayesian-flavored index policies in terms of regret. We also investigate their ability to attain an optimal rate in terms of Bayes risk. Some proofs are provided in the supplemental paper \cite{aosArxiv}.

\paragraph{Notation} Recall that $N_a(t) = \sum_{s=1}^t \ind_{(A_s=a)}$ is the number of draws from arm $a$ at the end of round $t$. Letting $\hat{\mu}_{a,s}=\frac{1}{s}\sum_{k=1}^s Y_{a,k}$ be the empirical mean of the first $s$ rewards from $a$, the empirical mean of arm $a$ after $t$ rounds of the bandit algorithm, $\hat{\mu}_a(t)$, satisfies $\hat{\mu}_a(t)=0$ if $N_a(t)=0$, $\hat{\mu}_a(t) = \hat{\mu}_{a,N_a(t)}$ otherwise.

\section{(Bayesian) exponential family bandit models\label{sec:Exponential}} In the rest of the paper, we consider the important class of \emph{exponential family bandit models}, in which the arms belong to a one-parameter canonical exponential family.

\subsection{Exponential family bandit model} 

A one-parameter canonical exponential family is a set $\cP$ of probability distributions, indexed by a real parameter  $\theta$ called the natural parameter, that is defined by
\[\cP = \{\nu_{\theta}, \theta \in \Theta : \nu_{\theta} \ \text{has a density} \ f_{\theta}(x)=\exp(\theta x - b(\theta)) \ \text{w.r.t} \ \xi \},\]
where $\Theta=(\theta^-,\theta^+) \subseteq \R$ is an open interval, $b$ a twice-differentiable and convex function (called the log-partition function) and $\xi$ a reference measure. Examples of such distributions include Bernoulli distributions, Gaussian distributions with known variance, Poisson distributions, or Gamma distributions with known shape parameter. 
 
% \begin{table}[h] 
% \centering
% %\hspace{-0.5cm} 
% \begin{tabular}{|c|c|c|c|c|}
% \hline
% Distribution & Density &  Mean $\mu$ & Parameter $\theta$ & $b(\theta)$ % & $d(\mu,\mu')$\\
% \\
% \hline
% Bernoulli $\cB(\lambda)$ & $\lambda^x(1-\lambda)^{1-x} \ind_{\{0,1\}}(x)$  & $\lambda$& $\log\frac{\lambda}{1-\lambda}$ & $\log(1+e^\theta)$ 
% %& $\mu \log\frac{\mu}{\mu'} + (1-\mu)\log\frac{1-\mu}{1-\mu'}$ \\
% \\
% \hline 
% Poisson $\cP(\lambda)$ & $\frac{\lambda^x}{x!}e^{-\lambda}\ind_{\N^*}(x)$ &  $\lambda$ &  $\log(\lambda)$ & $e^\theta$ 
% %& $\mu'-\mu + \mu\log\frac{\mu}{\mu'}$\\
% \\
% \hline
% Gaussian $\norm{\lambda}{\sigma^2}$   \vspace{-0.1cm}& $\frac{1}{\sqrt{2\pi\sigma^2}}e^{-\frac{(x-\lambda)^2}{2\sigma^2}}$ &  $\lambda$ & $\frac{\lambda}{\sigma^2}$ 
% & $\frac{\sigma^2\theta^2}{2}$ \\ %& $\frac{(\mu-\mu')^2}{2\sigma^2}$ \\
%  ($\sigma^2$ known) & & & & \\
% \hline
% %Exponential $\cE(\lambda)$& $\lambda e^{-\lambda x}\ind_{\R^+}(x)$ & $1/\lambda$ & $-\lambda$ & $-\log(-\theta)$  
% %& $\frac{\mu}{\mu'}-1-\log\frac{\mu}{\mu'}$\\
% %\\
% %\hline
% Gamma $\Gamma(k,\lambda)$ \vspace{-0.1cm}& $\frac{\lambda^k}{\Gamma(k)}x^{k-1} e^{-\lambda x}\ind_{\R^+}(x)$ & $k/\lambda$ & $-\lambda$ & $-k\log(-\theta)$  
% %& $k\left(\frac{\mu}{\mu'}-1-\log\frac{\mu}{\mu'}\right)$\\
% \\
% ($k$ known)&&&&\\
% \hline
% \end{tabular} 
% \caption{Examples of exponential families and associated divergence\label{table:FamilleExpo}.} %$\xi$ is the counting (resp. Lebesgue) measure in the first two (resp. last two) examples.}
% 
% %\vspace{-.3cm}
% \end{table}  

If $X\sim \nu_{\theta}$, it can be shown that $\bE[X]=\dot{b}(\theta)$ and $\mathrm{Var}[X]=\ddot{b}(\theta) >0$, where $\dot{b}$ (resp. $\ddot{b}$) is the derivative (resp. second derivative) of $b$ with respect to the natural parameter $\theta$. Thus there is a one-to-one mapping between the natural parameter $\theta$ and the mean $\mu= \dot{b}(\theta)$, and distributions in an exponential family can be alternatively parametrized by their mean. Letting $\mathrm{J} := \dot{b}(\Theta)$, for $\mu \in \mathrm{J}$ we denote by $\nu^\mu$ the distribution in $\cP$ that has mean $\mu$ : $\nu^\mu = \nu_{\dot{b}^{-1}(\mu)}$. The variance $\mathrm{V}(\mu)$ of the distribution $\nu^{\mu}$ is related to its mean in the following way:
\begin{equation}\mathrm{V}(\mu) = \ddot{b}(\dot{b}^{-1}(\mu)).\label{def:Variance}\end{equation}

In the sequel, we fix an exponential family $\cP$ and consider a bandit model $\nu^{\bm{\mu}} = (\nu^{\mu_1},\dots,\nu^{\mu_K})$, where $\nu^{\mu_a}$ belongs to $\cP$ and has mean $\mu_a$. When considering Bayesian bandit models, we restrict our attention to product prior distributions on $\bm \mu=(\mu_1,\dots,\mu_K)$, such that $\mu_a$ is drawn from a prior distribution on $\mathrm{J}=\dot{b}(\Theta)$ that has density $f_a$ with respect to the Lebesgue measure. We let $\pi_a^t$ be the posterior distribution on $\mu_a$ after the first $t$ rounds of the bandit game. With a slight abuse of notation, we will identify $\pi_a^t$ with its density, for which a more precise expression is provided in Section~\ref{subsec:ExpoBayesian}.

\subsection{Kullback-Leibler divergence and confidence intervals}

For distributions that belong to a one-parameter exponential family, the large deviation rate function has a simple and explicit form, featuring the Kullback-Leibler (KL) divergence, and one can build tight confidence intervals on their means. The KL-divergence between two distributions $\nu_\theta$ and $\nu_\lambda$ in an exponential family has a closed form expression as a function of the natural parameters $\theta$ and $\lambda$, given by 
\begin{equation}
 \Kl(\theta,\lambda):=\KL(\nu_{\theta},\nu_{\lambda})=\dot{b}(\theta)(\theta - \lambda) - b(\theta) + b(\lambda). \label{KLClose}
\end{equation}
We also introduce $\Kd(\mu,\mu')$ as the KL-divergence between the distributions of means $\mu$ and $\mu'$:
\[
 \mathrm{d}(\mu,\mu'):=\KL(\nu^\mu,\nu^{\mu'})=\Kl(\dot{b}^{-1}(\mu),\dot{b}^{-1}(\mu')). 
\]
Applying the Cram\'er-Chernoff method (see e.g. \cite{Boucheronal13CI}) in an exponential family yields an explicit deviation inequality featuring this divergence function: if $\hat{\mu}_{s}$ is the empirical mean of $s$ samples from $\nu^{\mu}$ and $x>\mu$, one has $\bP\left(\hat{\mu}_s > x \right)\leq \exp(-s d(x,\mu))$. This inequality can be used to build a confidence interval for $\mu$ based on a \emph{fixed number of observations} $s$. Inside a bandit algorithm, computing a confidence interval on the mean of an arm $a$ requires to take into account the \emph{random number of observations} $N_a(t)$ available at round $t$. Using a self-normalized deviation inequality (see \cite{KLUCBJournal} and references therein), one can show that, at any round $t$ of a bandit game, the kl-UCB index, defined as 
\begin{equation}u_a(t) := \sup \left\{q \in \JJ : N_a(t) d(\hat{\mu}_a(t),q) \leq \log(t\log^c(t))\right\},\label{def:KLUCBIndex}\end{equation}
where $c\geq 3$ is a real parameter, satisfies $\bP\left(u_a(t) > \mu_a\right)\gtrsim 1 - {1}/{(t\log^{c-2}t)}$ and is thus an upper confidence bound on $\mu_a$.
The \emph{exploration rate}, which is here $\log(t\log^c(t))$, controls the coverage probability of the interval. 

Closed-form expressions for the divergence function $d$ in the most common examples of exponential families are available (see \cite{KLUCBJournal}). Using the fact that $y \mapsto d(x,y)$ is increasing when $y>x$, an approximation of $u_a(t)$ can then be obtained using, for example, binary search. 

\subsection{Posterior distributions in Bayesian exponential family bandits\label{subsec:ExpoBayesian}} 
 
It is well-known that the posterior distribution on the mean of a distribution that belongs to an exponential family depends on two sufficient statistics: the number of observations and the empirical means of these observations. With $f_a$ the density of the prior distribution on $\mu_a$, introducing 
\[\pi_{a,n,x}(u) := \frac{\exp\left(n\left[\dot{b}^{-1}(u) x - b(\dot{b}^{-1}(u))\right]\right) f_a(u)}{\int_{\JJ}\exp\left(n\left[\dot{b}^{-1}(u) x - b(\dot{b}^{-1}(u))\right]\right) f_a(u)du} \ \ \ \text{for} \ \ \ \ u \in \JJ,\]
the density of the posterior distribution on $\mu_a$ after $t$ rounds of the bandit game can be written
\[\pi_a^t = \pi_{a,N_a(t),\hat{\mu}_a(t)}.\]
While our analysis holds for any choice of prior distribution, in practice one may want to exploit the existence of families of conjugate priors  (e.g. Beta distributions for Bernoulli rewards, Gaussian distributions for Gaussian rewards, Gamma distributions for Poisson rewards). With a prior distribution chosen in such a family, the associated posterior distribution is well-known and its quantiles are easy to compute, which is of particular interest for the Bayes-UCB algorithm, described in the next section.

% \begin{table}[h]
% \begin{center}
% \begin{tabular}{|c|c|c|}
% \hline
% Distribution  & Prior distribution   & Posterior distribution on $\mu$ after $n$ \\
% & on $\mu$ & observations with empirical mean $x$ \\
% \hline
% $\cB(\mu)$  & $\text{Beta}(a,b)$ & $\text{Beta}(a+nx,b+n(1-x))$ \\
% \hline
% $\cP(\mu)$ &  $\Gamma(c,d)$ & $\Gamma(c + nx, d + n)$ \\
% \hline
% $\norm{\mu}{\sigma^2}$  & $\norm{\mu_0}{m_0^{-1}}$ & 
% $\norm{\frac{m_0\mu_0 + nx\sigma^{-2}}{m_0 + n\sigma^{-2}}}{(m_0 + n\sigma^{-2})^{-1}}$ \\
% \hline
% %$\cE(1/\mu)$ &  $\text{Inv}\Gamma(c,d)$ & $\text{Inv}\Gamma(c + n,d + nx)$ \\
% %\hline
% $\Gamma(k,k/\mu)$  & $\text{Inv}\Gamma(c,d)$ & $\text{Inv}\Gamma(c + kn,d + knx)$\\
% \hline
% \end{tabular}
% \end{center}
% \caption{\label{table:PosteriorExp} Conjugate prior on the mean and associated posterior distributions.}
% \end{table}

Finally, we give below a rewriting of the posterior distribution that will be very useful in the sequel to obtain tight bounds on its tails.

\begin{lemma}\label{lem:PosteriorWriting}
\[\pi_{a,n,x}(u) = \frac{\exp(-nd(x,u)) f_a(u)}{\int_{\JJ}\exp(-nd(x,u)) f_a(u)du}, \ \ \text{for all} \ \  u \in \JJ.\]
\end{lemma}

\paragraph{Proof} Let $u \in\JJ$. One has
\begin{eqnarray*}
\pi_{a,n,x}(u) &=& \frac{\exp\left(n\left[\dot{b}^{-1}(u) x - b(\dot{b}^{-1}(u))\right]\right) f_a(u)}{\int_{\JJ}\exp\left(n\left[\dot{b}^{-1}(u) x - b(\dot{b}^{-1}(u))\right]\right) f_a(u)du} \times \frac{e^{-n\left[x\dot{b}^{-1}(x) - b(\dot{b}^{-1}(x))\right]}}{e^{-n\left[x\dot{b}^{-1}(x) - b(\dot{b}^{-1}(x))\right]}}\\
 & = & \frac{\exp\left(-n\left[x(\dot{b}^{-1}(x) -\dot{b}^{-1}(u))- b(\dot{b}^{-1}(x))  + b(\dot{b}^{-1}(u))\right]\right) f_a(u)}{\int_{\JJ}\exp\left(-n\left[x(\dot{b}^{-1}(x) -\dot{b}^{-1}(u))- b(\dot{b}^{-1}(x))  + b(\dot{b}^{-1}(u))\right]\right) f_a(u)du} \\
 & = & \frac{\exp(-nd(x,u)) f_a(u)}{\int_{\JJ}\exp(-nd(x,u)) f_a(u)du},
\end{eqnarray*}
 using the closed form expression \eqref{KLClose} and the fact that $\theta = \dot{b}^{-1}(\mu)$.

\section{Bayes-UCB: a simple and optimal Bayesian index policy\label{sec:BayesUCB}}

\subsection{Algorithm and main result}

The Bayes-UCB algorithm is an index policy that was introduced by \cite{AISTATS12} in the context of parametric bandit models. Given a prior distribution on the parameters of the arms, the index used for each arm is a well-chosen quantile of the (marginal) posterior distributions of its mean. 
For exponential family bandit models, given a product prior distribution on the means, the Bayes-UCB index is
\[{q}_a(t):= \ Q \left(1 - \frac{1}{t(\log t)^c} ; \pi_a^t\right) = Q \left(1 - \frac{1}{t(\log t)^c} ; \pi_{a,N_a(t),\hat{\mu}_a(t)}\right),\]
where $Q(\alpha;\pi)$ is the quantile of order $\alpha$ of the distribution $\pi$ (that is, $\bP_{X\sim \pi}(X \leq Q(\alpha;\pi))=\alpha$) and $c$ is a real parameter. In the particular case of bandit models with Gaussian arms, \cite{Reverdy14} have introduced a variant of Bayes-UCB with a slightly different tuning of the confidence level, under the name UCL (for Upper Credible Limit). 

While the efficiency of Bayes-UCB has been demonstrated even beyond bandit models with independent arms, regret bounds are available only in very limited cases. For Bernoulli bandit models asymptotic optimality is established by \cite{AISTATS12} when a uniform prior distribution on the mean of each arm is used. For Gaussian bandit models \cite{Reverdy14} give a logarithmic regret bound when an uniformative prior is used. In this section, we provide new finite-time regret bounds that hold in general exponential family bandit models, showing that a slight variant of Bayes-UCB is asymptotically optimal for a large class of prior distributions. 

We fix an exponential family, characterized by its log-partition function $b$ and the interval $\Theta=]\theta^-,\theta^+[$ of possible natural parameters. We let $\mu^- = \dot{b}(\theta^-)$ and $\mu^+ = \dot{b}(\theta^+)$ ($\mu^-$ may be equal to $-\infty$ and $\mu^+$ to $+\infty$). We analyze Bayes-UCB for exponential bandit models satisfying the following assumption. 

\begin{assumption}\label{assum:KnownBounds}
 There exists $\mu_0^->\mu^-$ and $\mu_0^+<\mu^+$ such that 
 $\forall a \in \{1,\dots,K\}, \ \ \mu_0^- \leq \mu_a \leq \mu_0^+.$
\end{assumption}

For Poisson or Exponential distributions, this assumption requires that the means of all arms are different from zero, while they should be included in $]0,1[$ for Bernoulli distributions. We now introduce a regularized version of the Bayes-UCB index that relies on the knowledge of $\mu_0^-$ and $\mu_0^+$, as   
\begin{equation}\overline{q}_a(t):= \ Q \left(1 - \frac{1}{t(\log t)^c} ; \pi_{a,N_a(t),\bar{\mu}_a(t)}\right),\label{index:VariantBayes}\end{equation}
where $\bar{\mu}_a(t) = \min\left(\max(\hat{\mu}_a(t),\mu_0^-),\mu_0^+\right)$. Note that $\mu_0^-$ and $\mu_0^+$ can be chosen arbitrarily close to $\mu^-$ and $\mu^+$ respectively, in which case $\overline{q}_a(t)$ often coincides with the original Bayes-UCB index $q_a(t)$.

\begin{theorem}\label{thm:BayesUCB} Let $\nu^{\bm{\mu}}$ be an exponential bandit model satisfying Assumption~\ref{assum:KnownBounds}. 
 Assume that for all $a$, $\pi_a^0$ has a density $f_a$ with respect to the Lebesgue measure such that $f_a(u)>0$ for all $u \in \mathrm{J} = \dot{b}(\Theta)$. Let $c\geq 7$. The algorithm that draws each arm once and for $t\geq K$ selects at time $t+1$ 
 \[A_{t+1} = \underset{a}{\emph{argmax}} \ \overline{q}_a(t),\]
 with $\overline{q}_a(t)$ defined in \eqref{index:VariantBayes} satisfies, for all $\varepsilon>0$,
 \[\forall a \neq a^*, \ \ {\bE[N_a(T)]}\leq \frac{1+\varepsilon}{d(\mu_a,\mu^*)}\log(T) + o_{\varepsilon}\left(\log(T)\right).\]
\end{theorem}

From Theorem~\ref{thm:BayesUCB}, taking the $\limsup$ and letting $\epsilon$ go to zero show that (this slight variant of) Bayes-UCB satisfies   
 \[\forall a \neq a^*, \ \ \limsup_{T\rightarrow \infty} \frac{\bE[N_a(T)]}{\log(T)} \leq \frac{1}{d(\mu_a,\mu^*)}.\]
Thus this index policy is asymptotically optimal, as it matches Lai and Robbins' lower bound \eqref{equ:LR}. As we shall see in Section~\ref{sec:Experiments}, from a practical point of view Bayes-UCB outperforms kl-UCB and performs similarly (sometimes slightly better, sometimes slightly worse) as Thompson Sampling, another popular Bayesian algorithm that we now discuss.  

\subsection{Posterior quantiles versus posterior samples}

Over the past few years, another Bayesian algorithm, Thompson Sampling, has become increasingly popular for its good empirical performance, and we explain how Bayes-UCB is related to this alternative, randomized, Bayesian approach. 

The Thompson Sampling algorithm, that draws each arm according to its posterior probability of being optimal, was introduced in 1933 as the very first bandit algorithm \cite{Thompson33} and re-discovered recently for its good empirical performance \cite{Scott10,LiChapelle11}. Thompson Sampling can be implemented in virtually any Bayesian bandit model in which one can sample the posterior distribution, by drawing \emph{one} sample from the posterior on each arm and selecting the arm that yields the largest sample. In any such case, Bayes-UCB can be implemented as well and may appear as a more robust alternative as the quantiles can be estimated based on \emph{several} samples in case there is no efficient algorithm to compute them.    

Our experiments of Section~\ref{sec:Experiments} show that Bayes-UCB as well as the other Bayesian-flavored index policies presented in Section~\ref{sec:KLUCBPlus} are competitive with Thompson Sampling in general one-dimensional exponential families. Compared to Bayes-UCB, the theoretical understanding of Thompson Sampling is more limited: this algorithm is known to be asymptotically optimal in exponential family bandit models, yet only for specific choices of prior distributions \cite{ALT12,AGAISTAT13,NIPS13}. 

In more complex bandit models, there are situations in which Bayes-UCB is indeed used over Thompson Sampling. When there is a potentially infinite number of arms and the mean reward function is assumed to be drawn from a Gaussian Process, the GP-UCB of \cite{SrinivasGPUCB}, that coincides with Bayes-UCB, is very popular in the Bayesian optimization community \cite{Brochu10Tuto}.

\subsection{Tail bounds for posterior distributions}

Just like the analysis of \cite{AISTATS12}, the analysis of Bayes-UCB that we give in the next section relies on tight bounds on the tails of posterior distributions that permit to control quantiles. These bounds are expressed with the Kullback-Leibler divergence function $d$. Therefore, an additional tool in the proof is the control of the deviations of the empirical mean rewards from the true mean reward, measured with this divergence function, which follows from the work of \cite{KLUCBJournal}. 

In the particular case of Bernoulli bandit models, Bayes-UCB uses quantiles of Beta posterior distributions. In that case a specific argument, namely the fact that $\text{Beta}(a,b)$ is the distribution of the $a$-th order statistic among $a+b-1$ uniform random variables, relates a Beta distribution (and its tails) to a Binomial distribution (and its tails). This `Beta-Binomial trick' is also used extensively in the analysis of Thompson Sampling for Bernoulli bandits proposed by \cite{AGCOLT12,ALT12,AGAISTAT13}. Note that this argument can only be used for Beta distributions with integer parameters, which rules out many possible prior distributions. The analysis of \cite{Reverdy14} in the Gaussian case also relies on specific tails bounds for the Gaussian posterior distributions.  For exponential family bandit models, an upper bound on the tail of the posterior distribution was obtained by \cite{NIPS13} using the Jeffrey's prior.  

Lemma~\ref{lem:DeterministicULB} below present more general results that hold for any class of exponential family bandit models and any prior distribution with a density that is positive on $\JJ=\dot{b}(\Theta)$. 
 For such (proper) prior distributions, we give deterministic upper and lower bounds on the corresponding posterior probabilities  $\pi_{a,n,x}([v,\mu^+[)$. Compared to the result of \cite{NIPS13}, which is not presented in this deterministic way, Lemma~\ref{lem:DeterministicULB} is based on a different rewriting of the posterior distribution, given in Lemma \ref{lem:PosteriorWriting}.

\begin{lemma}\label{lem:DeterministicULB} 
Let $\mu_0^-,\mu_0^+$ be defined in Assumption~\ref{assum:KnownBounds}. 
\begin{enumerate}
\item There exist two positive constants $A$ and $B$ such that for all $x,v$ that satisfy 
$\mu_0^-<x<v<\mu_0^+$, for all $n\geq 1$, for all $a\in \{1,\dots,K\}$,
\[ {A}{n^{-1}}e^{-nd(x,v)} \leq \pi_{a,n,x}([v,\mu^+[) \leq B \sqrt{n} e^{-nd(x,v)}.\]
\item There exists a constant $C$ such that for all $x,v$ that satisfy 
$\mu_0^-<v\leq x <\mu_0^+$, for all $n\geq 1$, for all $a\in \{1,\dots,K\}$,
\[\pi_{a,n,x}([v,\mu^+[) \geq \frac{C}{\sqrt{n}}.\]
\end{enumerate}
The constants $A,B,C$ depend on $\mu_0^-$,$\mu_0^+$, $b$ and the prior densities. 
\end{lemma}

This result permits in particular to show that the quantile $\overline{q}_a(t)$ defined in~\eqref{index:VariantBayes} satisfies $\underline{\mathrm{U}}_a(t) \leq \overline{q}_a(t) \leq \overline{\mathrm{U}}_a(t)$, with 
\begin{eqnarray*}
\underline{\mathrm{U}}_a(t) & = & \sup \big\{ q < \mu_0^+ : N_a(t) d( \overline{\mu}_a(t), q ) \leq \log\left((At\log^c(t))/N_a(t)\right)\big\}, \\ 
\overline{\mathrm{U}}_a(t) & = & \sup \left\{ q < \mu_0^+ : N_a(t) d( \overline{\mu}_a(t), q ) \leq \log\left(Bt\log^c(t)\sqrt{N_a(t)}\right)\right\}. 
\end{eqnarray*}
Hence, despite their Bayesian nature, the indices used in Bayes-UCB are strongly related to frequentist kl-UCB type indices. However, compared to the index $u_a(t)$ defined in \eqref{def:KLUCBIndex}, the exploration rate that appears in  $\underline{\mathrm{U}}_a(t)$ and $\overline{\mathrm{U}}_a(t)$  also features the current number of draws $N_a(t)$. Lai gives in \cite{Lai87} an asymptotic analysis of any index strategy of the above form with an exploration function $g(T/N_a(t))$, where $g(t)\sim \log(t)$ when $t$ goes to infinity. Yet neither $\underline{\mathrm{U}}_a(t)$ nor $\overline{\mathrm{U}}_a(t)$ are not exactly of that form, and we propose below a finite-time analysis that relies on new, non-asymptotic, tools.  

\subsection{Finite-time analysis}

We give here the proof of Theorem~\ref{thm:BayesUCB}. To ease the notation, assume that arm 1 is an optimal arm, and let $a$ be a suboptimal arm. 
\[\bE[N_a(T)] = \bE\left[\sum_{t=0}^{T-1} \ind_{(A_{t+1}=a)}\right] = 1 + \bE\left[\sum_{t=K}^{T-1} \ind_{(A_{t+1}=a)}\right].\]
We introduce a truncated version of the KL-divergence, $d^+(x,y) := d(x,y) \ind_{(x<y)}$ and let $g_t$ be a decreasing sequence to be specified later.

Using that, by definition of the algorithm, if $a$ is played at round $t+1$, it holds in particular that $\overline{q}_a(t)\geq \overline{q}_1(t)$, one has
\begin{eqnarray*}
(A_{t+1} = a )  &\subseteq&  \left(\mu_1 - g_t \geq \overline{q}_1(t)\right) \bigcup \left(\mu_1 - g_t \leq \bar{q}_1(t), A_{t+1}=a\right) \\ 
 &\subseteq&  \left(\mu_1 - g_t \geq \overline{q}_1(t)\right) \bigcup \left(\mu_1 - g_t \leq \bar{q}_a(t), A_{t+1}=a\right).
\end{eqnarray*}
This yields
\[\bE[N_a(T)] \leq 1 + \!{\sum_{t=K}^{T-1} \bP\left(\mu_1 - g_t \geq \bar{q}_1(t)\right)} + 
{\sum_{t=K}^{T-1} \bP\left(\mu_1 - g_t \leq \bar{q}_a(t),A_{t+1}=a\right)}.\]

The posterior bounds established in Lemma~\ref{lem:DeterministicULB} permit to further upper bound the two sums in the right-hand side of the above inequality. With $C$ defined in  Lemma~\ref{lem:DeterministicULB}, we introduce $t_0$, defined by 
\[t\geq t_0 \ \ \Rightarrow \ \  \left( \mu_1-g_t \geq \mu_0^-  \ \ \text{and} \ \  C^2t\log(t)^{2c} > 1\right).\] 

On the one hand, for $t\geq t_0$, 
\begin{align*}
 &\left(\mu_1 - g_t \geq \bar{q}_1(t)\right)  =  \left(\pi_{1,N_1(t),\bar{\mu}_1(t)}([\mu_1-g_t, \mu^+[) \leq \frac{1}{t\log^c t}\right) \\
 & \hspace{2.8cm}=  \left(\pi_{1,N_1(t),\bar{\mu}_1(t)}([\mu_1-g_t, \mu^+[) \leq \frac{1}{t\log^c t}, \bar{\mu}_1(t) \leq \mu_1 - g_t\right),
\end{align*}
since by the lower bound in the second statement of Lemma~\ref{lem:DeterministicULB},
\begin{align*}
&\left(\pi_{1,N_1(t),\bar{\mu}_1(t)}([\mu_1-g_t, \mu^+[) \leq \frac{1}{t\log^c t}, \bar{\mu}_1(t) \geq \mu_1 - g_t\right)\\
&\hspace{1cm}\subset \left(\frac{C}{\sqrt{N_1(t)}} \leq \frac{1}{t\log^c t}\right) \subset  \left(N_1(t) \geq C^2t^2\log^{2c} t\right) 
%& \hspace{1cm}
\subset \left(N_1(t) > t\right)=\emptyset.
\end{align*}
Now using the lower bound in the first statement of Lemma~\ref{lem:DeterministicULB},
\begin{eqnarray*}
 \left(\mu_1 - g_t \geq \bar{q}_1(t)\right) & \subseteq & \left(\frac{Ae^{-N_1(t)d(\bar{\mu}_1(t),\mu_1-g_t)}}{N_1(t)} \leq \frac{1}{t\log^c t}, \bar{\mu}_1(t) \leq \mu_1 - g_t\right) \\
 & \subset & \left(N_1(t) d^+(\hat{\mu}_1(t),\mu_1-g_t) \geq \log\left(\frac{At\log^c t}{N_1(t)}\right)\right).
\end{eqnarray*}

On the other hand, 
\begin{align}
& \sum_{t=K}^{T-1} \bP\left(\mu_1 - g_t \leq \bar{q}_a(t),A_{t+1}=a\right) \nonumber \\
& =  \sum_{t=K}^{T-1} \bP\left(\pi_{a,N_a(t),\bar{\mu}_a(t)} ([\mu_1-g_t, \mu^+[) \geq \frac{1}{t\log^c t},A_{t+1}=a\right) \nonumber \\
& \leq \sum_{t=K}^{T-1} \bP\left(\bar{\mu}_a(t) < \mu_1 - g_t, \pi_{a,N_a(t),\bar{\mu}_a(t)} ([\mu_1-g_t, \mu^+[) \geq \frac{1}{t\log^c t}, A_{t+1}=a\right)\label{TBC2} \\
& \hspace{2cm} + \sum_{t=K}^{T-1}\bP\left(\bar{\mu}_a(t) \geq \mu_1 - g_t, A_{t+1} = a\right).\nonumber
\end{align}
Using Lemma~\ref{lem:DeterministicULB}, the first sum in \eqref{TBC2} is upper bounded by
\begin{align*}
& \sum_{t=K}^{T-1} \bP\left(B\sqrt{N_a(t)}e^{-N_a(t) d^+(\bar{\mu}_a(t),\mu_1-g_t)}\geq \frac{1}{t\log^c t}, A_{t+1}=a\right) \\
& \hspace{0.5cm}\leq \sum_{t=K}^{T-1} \sum_{s=1}^t\bP\left(B\sqrt{s}e^{-s d^+(\bar{\mu}_{a,s},\mu_1-g_t)}\geq \frac{1}{t\log^c t}, N_a(t) = s, A_{t+1}=a\right) \\
& \hspace{0.5cm}\leq \sum_{t=K}^{T-1} \sum_{s=1}^t\bP\left(sd^+(\bar{\mu}_{a,s},\mu_1-g_s) \leq \log(T\log^c T) + \log(B) + \frac{1}{2}\log s , \right. \\ 
& \hspace{8.5cm} N_a(t) = s, A_{t+1}=a\Big) \\
& \hspace{0.5cm}\leq \sum_{s=1}^T\bP\left(sd^+(\bar{\mu}_{a,s},\mu_1-g_s) \leq \log T + c \log\log T + \log(B) + \frac{1}{2}\log s\right) \\
& \hspace{0.5cm}\leq \sum_{s=1}^T\bP\left(sd^+(\hat{\mu}_{a,s},\mu_1-g_s) \leq \log T + c \log\log T + \log(B) + \frac{1}{2}\log s\right) \\
& \hspace{1cm}+ \sum_{s=1}^T\bP(\hat{\mu}_{a,s} < \mu_0^-). 
\end{align*}
To third inequality follows from exchanging the sums over $s$ and $t$ and using that $\sum_{t=1}^N \ind_{(N_a(t) = s)\cap(A_{t+1}=a)}$ is smaller than 1 for all $s$. The last inequality uses that if $\hat{\mu}_{a,s} \geq \mu_0$, $\overline{\mu}_{a,s} \leq \hat{\mu}_{a,s}$ and $d^+(\overline{\mu}_{a,s},\mu_1-g_s) \geq d^+(\hat{\mu}_{a,s},\mu_1 - g_s)$. Then by Chernoff inequality, 
\[\sum_{s=1}^T\bP(\hat{\mu}_{a,s} < \mu_0^-) \leq \sum_{s=1}^\infty \exp(-sd(\mu_0^-,\mu_a)) = \frac{1}{1-e^{-d(\mu_0^-,\mu_a)}}.
\]
Still using Chernoff inequality, the second sum in \eqref{TBC2} is upper bounded by  
\begin{align*}
& \sum_{t=K}^{T-1}\bP\left(\hat{\mu}_a(t) \geq \mu_1 - g_t, A_{t+1} = a\right) \leq \sum_{t=K}^{T-1} \bP\left(\hat{\mu}_a(t) \geq \mu_1 - g_{N_a(t)} , A_{t+1}=a\right) \\ 
&\leq \sum_{t=K}^{T-1}\sum_{s=1}^t\bP\left(\hat{\mu}_{a,s} \geq \mu_1 - g_s, N_a(t)=s, A_{t+1} = a\right)  \\
& \leq  \sum_{s=1}^T \bP\left(\hat{\mu}_{a,s} \geq \mu_1 - g_s\right)  \leq \sum_{s=1}^\infty \exp(-sd(\mu_1-g_s,\mu_a)):= N_0 < +\infty.
\end{align*}

Putting things together, we showed that there exists some constant $N = \max (t_0,N_0+(1-e^{-d(\mu_0^-,\mu_a)})^{-1})+1$ such that 
\begin{align*}
& \bE[N_a(T)] \leq N + \underbrace{\sum_{t=K}^{T-1} \bP\left(N_1(t) d^+(\hat{\mu}_1(t),\mu_1-g_t) \geq \log\left(\frac{At\log^c t}{N_1(t)}\right)\right)}_{T_1} \\
& + \underbrace{\sum_{s=1}^T\bP\left(sd^+(\hat{\mu}_{a,s},\mu_1-g_s) \leq \log T + c \log\log T + \log(B) + \frac{1}{2}\log s\right)}_{T_2}
\end{align*}

Term $T_1$ is shown below to be of order $o(\log(T))$, as $\hat{\mu}_1(t)$ cannot be too far from $\mu_1 - g_t$. Note however that the deviation is expressed with $\log(t/N_1(t))$ in place of the traditional $\log(t)$, which makes the proof of Lemma~\ref{lem:TermA} more intricate. In particular, Lemma~\ref{lem:TermA} applies to a specific sequence $(g_t)$ defined therein, and a similar result could not be obtained for the choice $g_t=0$, unlike Lemma~\ref{lem:TermB} below. 

\begin{lemma}\label{lem:TermA} Let $g_t$ be such that $d(\mu_1-g_t,\mu_1)=\frac{1}{\log(t)}$. If $c\geq 7$, for all $A$, if $t$ is larger than $\exp(\max(\sqrt{3},A^{-1/7}))$,  
\begin{align*}
& \bP\left(N_1(t) d^+(\hat{\mu}_1(t),\mu_1-g_t) \geq \log \frac{At\log^c t}{N_1(t)}\right)   \\
&\hspace{1cm}
\leq e \left( \frac{1}{At\log t} + \frac{3\log\log t + \log A}{At\log^2 t} + \frac{1}{A t \log^3 t}\right) + \frac{1}{t^2}. 
\end{align*}
\end{lemma}

From Lemma~\ref{lem:TermA}, one has 
\begin{align*}
  (T_1) & \leq e \sum_{t=K}^{T-1}\frac{\log^2 t + 3(\log t)\log\log(t) +\log A \log t + 1}{At(\log^3 t)}  + \sum_{t=K}^{T-1}\frac{1}{t^2}\\
  & \leq  \frac{e}{A}\left(2+\frac{3}{e}+\frac{\log A }{\log K}\right)\sum_{t=K}^{T-1}\frac{1}{t\log(t)} + \frac{\pi^2}{6} \\ &\leq  \frac{e}{A}\left(2+\frac{3}{e}+\frac{\log A}{\log K}\right)\log \log T + \frac{\pi^2}{6}.
\end{align*}

The following lemma permits to give an upper bound on Term T2. 

\begin{lemma}\label{lem:TermB} Let $f,g,h$ be three functions such that 
\[f(s) \underset{s \rightarrow \infty}{\longrightarrow} \infty, \ \ \ \ g(s) \underset{s \rightarrow \infty}{\longrightarrow} 0 \ \ \ \text{and} \ \ \ \frac{h(s)}{s}\underset{s\rightarrow \infty}{\longrightarrow} 0,\]
with $g$ and $s \mapsto h(s)/s$ non-increasing for $s$ large enough.

For all $\varepsilon>0$ there exists a (problem-dependent) constant $N_a(\varepsilon)$ such that for all $T \geq N_a(\varepsilon)$, 
\begin{align*}
&\sum_{s=1}^T\bP\left(sd^+(\hat{\mu}_{a,s},\mu_1-g(s)) \leq  f(T) + h(s)\right)  \\
& \hspace{0.6cm}\leq  \frac{1+\varepsilon}{d(\mu_a,\mu_1)}f(T) + \sqrt{f(T)} \sqrt{\frac{8\mathrm{V}_a^2\pi(1+\varepsilon)^3d'(\mu_a,\mu_1)^2}{d(\mu_a,\mu_1)^3}} \\ 
&\hspace{0.8cm}+\  8(1+\varepsilon)^2 \mathrm{V}_a^2\left(\frac{d'(\mu_a,\mu_1)}{d(\mu_a,\mu_1)}\right)^2 \frac{1}{1-e^{-d(\mu_0^-,\mu_a)}} + 1,
\end{align*}
with $\mathrm{V}_a = \sup_{\mu \in [\mu_a,\mu_1]} \mathrm{V}(\mu)$, where the variance function is defined in \eqref{def:Variance}.
\end{lemma}

Let $\varepsilon>0$. Using Lemma~\ref{lem:TermB}, with $f(s)=\log(s)+c\log\log(s) + \log(B)$, $g(s)=g_s$ defined in Lemma~\ref{lem:TermA} and $h(s)=\frac{1}{2}\log(s)$, there exists problem dependent constants $C_0$ and $D_0(\varepsilon)$ such that  
\begin{align*}
  (T_2) & \leq \frac{1+\varepsilon}{d(\mu_a,\mu_1)}(\log T + c \log\log T) + C_0\sqrt{\log T + c \log\log T} + D_0(\varepsilon).
\end{align*}
Putting together the upper bounds on (T1) and (T2) yields the conclusion: for all $\varepsilon>0$, 
\[\bE[N_a(T)] \leq \frac{1+\varepsilon}{d(\mu_a,\mu^*)}\log(T) + O_{\varepsilon}(\sqrt{\log(T)}).\]

\section{A Bayesian insight on alternative exploration rates}\label{sec:KLUCBPlus}

The kl-UCB index of an arm, $u_a(t)$, introduced in \eqref{def:KLUCBIndex}, uses the exploration rate $\log(t\log^c(t))$, that does not depend on arm $a$. Some alternatives to this universal exploration rate have been suggested in the literature, and we formally introduce two variants of kl-UCB, called kl-UCB$^+$ and kl-UCB-H$^+$ using an exploration rate that decreases with the number of draws of arm $a$. The tools developed for the analysis of Bayes-UCB allow us to prove the asymptotic optimality of both algorithms. We then show that the Bayesian literature on the multi-armed bandit problem provides a natural justification for these algorithms, that are related to approximations of the Bayesian optimal optimal solution or the Gittins indices.

\subsection{The kl-UCB$^+$ and kl-UCB-H$^+$ algorithms}\label{subsec:KLUCBPlus}
  
We introduce in Definition~\ref{def:AlternativeIndices} two new index policies, and prove their asymptotic optimality. The indices indices $u_a^{H,+}(t)$ and $u_a^+(t)$ both rely on  an exploration rate that decreases with the number of plays of arm $a$. kl-UCB-H$^+$ additionally requires the knowledge of the horizon $T$. In practice, both algorithms outperform kl-UCB, as can be seen in Section~\ref{sec:Experiments}. 

\begin{definition}\label{def:AlternativeIndices} Let $c\geq 0$. We define kl-UCB-H$^+$ and kl-UCB$^+$ with parameter $c\geq 0$ as the index policies respectively based on the indices 
\begin{eqnarray}
u_a^{H,+}(t) & = & \sup  \left\{ q  : N_a(t) d(\hat{\mu}_a(t),q) \leq \log\left(\frac{T\log^c T}{N_a(t)}\right)\right\},\label{index:KLUCBHPlus}\\
u_a^+(t) & = & \sup  \left\{ q : N_a(t) d(\hat{\mu}_a(t),q) \leq \log \left(\frac{t\log^c t}{N_a(t)}\right)\right\}.\label{index:KLUCBPlus}
\end{eqnarray}
\end{definition}

A key step in the analysis of Bayes-UCB is the control of the probability of the event
\[\left(N_1(t) d^+(\hat{\mu}_1(t),\mu_1-g_t) \geq \log \left(\frac{At\log^c t}{N_1(t)}\right)\right),\]
in which an exploration rate of order $\log(t/N_1(t))$ appears. This control is obtained in Lemma~\ref{lem:TermA} which
can also be used to analyze the kl-UCB-H$^+$ and kl-UCB$^+$ algorithms, that are based on such alternative exploration rates. The following theorem proves the asymptotic optimality of these two index policies. The proof is provided in Appendix~\ref{proof:FiniteTime}. 

\begin{theorem}\label{thm:KLvariants} Let $c\geq 7$. Each of the index policy associated to the indices defined by \eqref{index:KLUCBPlus} and \eqref{index:KLUCBHPlus} satisfies, for all $\varepsilon>0$, 
\[\bE[N_a(T)] \leq \frac{1+\varepsilon}{d(\mu_a,\mu^*)}\log(T) + O_{\varepsilon}(\sqrt{\log(T)}).\]
\end{theorem}

The use of alternative exploration rates in UCB-type algorithms has appeared before in the bandit literature. For example the MOSS algorithm \cite{Audibertal10MOSS}, based on the index
\[\hat{\mu}_a(t) + \sqrt{\frac{\log\left({T}/({KN_a(t)})\right)}{N_a(t)}},\]
is designed to be optimal in a minimax sense for bandit models with sub-gaussian rewards: the algorithm achieves a $O(\sqrt{KT})$ distribution-independent upper bound on the regret. Besides, it was already noted by \cite{AOKLUCB} that the use of the exploration rate $\log(t/N_a(t))$ in place of $\log(t)$ in the kl-UCB algorithm leads to better empirical performance. 
In this paper, additionally to proving the asymptotic optimality of these approaches, we now provide a new insight on the use of such alternative exploration rates by relating the kl-UCB-H$^+$ algorithm to other Bayesian policies.

\subsection{Bayesian optimal solution and Gittins indices\label{sec:Gittins}}

The alternative exploration rate discussed in Section~\ref{subsec:KLUCBPlus} happens to be related to two other Bayesian strategies for the multi-armed bandit problem: the Bayesian optimal solution and the Finite-Horizon Gittins index policy, that we present here.

In a Bayesian framework, the interaction of an agent with a multi-armed bandit can be modeled by a Markov Decision Process (MDP) in which the state $\Pi_t$ is the current posterior distribution over the parameter of the arms. In exponential bandit models, the posterior over $\bm \mu$ is $\Pi_t = {{\bigotimes}} \pi_a^t$. There are $K$ possible actions and when action $A_t$ is chosen in state $\Pi_t$, the observed reward $X_t$ is a sample from arm $A_t$, that satisfies, conditionally to the past, $X_t  \sim  \nu^{\mu}$ and $\mu  \sim  \Pi_t(A_t)$. The new state is $\Pi^{t+1} =  {{\bigotimes}} \pi_a^{t+1}$ with $\pi_a^{t+1}=\pi_a^t$ for all $a\neq A_t$ and the density of $\pi_{A_t}^{t+1}$ gets updated according to
\[
\pi_{A_t}^{t+1}(u) \propto \exp(-(\dot{b}^{-1}(u) X_t - b(\dot{b}^{-1}(u))))\pi_{A_t}^t(u).
\]
Bayes risk minimization, or reward maximization under the Bayesian probabilistic model, is equivalent to solving this MDP for the finite-horizon criterion, which boils down to finding a strategy of the form $A_t = g(\Pi_t)$ for some deterministic function $g$, that maximizes  
\begin{equation}\bE^{\Pi} \left[\sum_{t=1}^T X_t^g\right],\label{equ:maxReward}\end{equation}
where $(X_t^g)_t$ is the sequence of rewards obtained under policy $g$.
From the theory of MDPs (see e.g., \cite{Puterman94MDP}), the optimal policy is solution of dynamic programming equations and can be computed by induction. However, due to the very large, if not infinite, state space (the set of possible posterior distributions over $\bm \mu$), the computation is often intractable.

In a slightly different setting, Gittins proved in 1979 \cite{Gittins79} that the apparently intractable optimal policy reduces to an index policy, with corresponding indices later called the \emph{Gittins indices}. He considers the discounted Bayesian multi-armed bandit problem, in which the goal is to find a policy $g$ that minimizes
\[\bE^{\Pi} \left[\sum_{t=1}^\infty \alpha^{t-1}X_t^g\right],\]
for some discount parameter $\alpha \in ]0,1[$. Interestingly, it was proved in \cite{BerryFristedt85} that the discount is necessary for this reduction to hold: in particular, the policy maximizing \eqref{equ:maxReward} is $\emph{not}$ an index policy.  
However, the notion of Gittins indices is a powerful concept that can also be defined in a finite horizon multi-armed bandit. The Finite-Horizon Gittins index of an arm depends on the current posterior distribution on its mean ($\pi=\pi_a^t$) and on the remaining time to play ($r = T-t$). It can be interpreted as the price worth paying for playing an arm with posterior $\pi$ at most $r$ times. Indeed, for $\lambda>0$ consider the following game, called $\cC_\lambda$, in which a player can either pay $\lambda$ and draw the arm to receive a sample $Y_t$, which results in a reward $Y_t-\lambda$, or stop playing, which yields no reward. As precisely defined below, the Gittins index is the critical value of $\lambda$ for which the optimal policy in $\cC_\lambda$ is to stop playing the arm from the beginning. This definition transposes to the non-discounted case one of the equivalent definitions of the discounted Gittins index that can be found in \cite{GittinsBook11}.

\begin{definition}\label{def:FHG} The Finite-Horizon Gittins index for a current posterior $\pi$ and remaining time $r$ is $G(\pi,r) = \inf \{ \lambda \in \R : V_\lambda^*(\pi,r)  = 0\},$ with
\[V^*_\lambda(\pi,r) = \sup_{0\leq \tau \leq r} \bE_{\substack{Y_t \overset{\text{i.i.d}}{\sim} \nu^\mu \\ \mu \sim \pi}}\left[\sum_{t=1}^\tau (Y_t- \lambda)\right],\]
where the supremum is taken over all stopping time $\tau$ smaller than $r$ a.s., with the convention $\sum_{t=1}^{0} \cdot =0$.
\end{definition}

\begin{figure}[h]
\centering
\includegraphics[height=5cm]{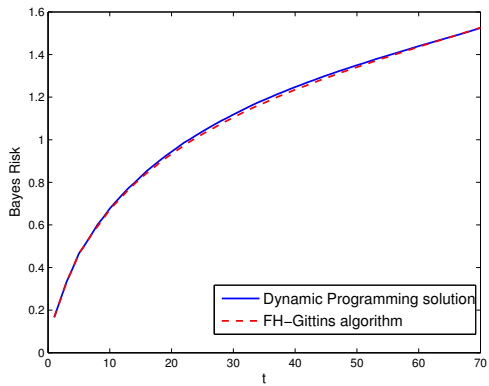} 
\caption{\label{fig:GittinsOPT} Bayes risk of the optimal strategy (blue) and FH-Gittins (dashed red) estimated using $N=10^6$ replications of a bandit game, for which the means are drawn from $\cU([0,1])$}
\end{figure}

Computing the FH-Gittins indices requires to compute $V_\lambda^*(\pi,r)$ for several values of $\lambda$ in order to find the critical value (using, e.g., binary search). Each computation requires solving a MDP, but on a smaller state space: the possible posterior distributions on the mean of a single arm. Hence the FH-Gittins algorithm, that is the index policy based on the Finite-Horizon Gittins indices, 
\[A_{t+1} = \underset{a = 1,\dots,K}{\text{argmax}} \ G(\pi_a^t,T-t),\]
is a more practical algorithm than the Bayesian optimal solution. Although FH-Gittins does not coincide with the Bayesian optimal solution, we believe it is a good approximation. This is supported by simulations performed in a two-armed Bernoulli bandit problem, for which we compute the Bayes risk of the optimal strategy and that of the FH-Gittins algorithm up to horizon $T=70$, as presented in Figure~\ref{fig:GittinsOPT}. For small horizons, \cite{GinebraClayton99} propose a comparison of different algorithms with the Bayesian optimal solution and similarly notice that the Bayes risk of FH-Gittins (called $\Lambda$-strategy) 
is very close to the optimal value, for various choices of prior and horizons. 

Compared to a simple index policy like Bayes-UCB, the computational cost of the FH-Gittins algorithm (not to mention that of the Bayesian optimal strategy) is still very high. In particular, the complexity of these two approaches grows dramatically when the horizon $T$ increases, which motivates some approximations that have been proposed for large horizons, described in the next sections. 

However, when the FH-Gittins algorithm is efficiently implementable (that is, for relatively small horizons), we would like to advocate its use for minimizing the frequentist regret. Indeed our experiments of Section~\ref{sec:Experiments} report good empirical performance in Bernoulli bandit models. In this particular case, using a uniform prior on the means, the set of (Beta) posterior is parametrized by two integers (the number of zeros and  ones observed so far), and we could implement FH-Gittins up to horizon $T=1000$. An  efficient implementation of FH-Gittins for Gaussian bandits, up to horizon $T=10000$, has been recently given by  \cite{Tor15Gittins}. More generally, finding efficient methods to compute Finite-Horizon Gittins indices is still an area of investigation \cite{NinoMora11Finite}. Interestingly, \cite{Tor15Gittins} provides the first theoretical elements supporting the use of FH-Gittins for regret minimization, by giving the first logarithmic upper bound on its regret in the particular case of Gaussian bandit models. However, the asymptotic optimality of this algorithm for Gaussian bandits and more general models remains a conjecture.

\subsection{Approximation of the Bayesian optimal solution}

In the paper \cite{Lai87}, Lai shows that, in exponential family bandit models, the Bayes risk of \emph{any} strategy is asymptotically lower bounded by $C_0(\pi)\log^2(T)$, when $C_0(\pi)$ is a prior-dependent constant. He also provides matching strategies, which implies in particular that the Bayes risk of the Bayesian optimal solution is of order $\log^2(T)$. Any strategy matching this lower bound can be viewed as an asymptotic approximation of the Bayesian optimal solution. 

In the particular case of product prior distributions, we provide in Theorem~\ref{thm:LRB} a Bayes risk lower bound that is slightly more general than Lai's result in the sense that it does not require the prior distribution on the natural parameter of each arm to have a compact support. The proof of this result, provided in Appendix~\ref{proof:LRB}, follows however closely that of \cite{Lai87}. The lower bound is expressed in terms of the prior distribution on the natural parameters $\bm \theta = (\theta_1,\dots,\theta_K)$ of the arms, with the following notation.
 For $a=1,\dots,K$, we let $\thetaa=(\theta_1,\dots,\theta_{a-1},\theta_{a+1},\dots,\theta_K)$ be the vector of $\Theta^{K-1}$ that consists of all components of $\thetaB$ except component number $a$. We let $\theta_a^* = \max_{i \neq a} \theta_i$, so that $\theta_a^*$ only depends on $\thetaa$. 
 
\begin{theorem}\label{thm:LRB} Let $H$ be a prior distribution on $\Theta^K$ that has a product form, such that each marginal has a density $h_a$ with respect to the Lebesgue measure $\lambda$ that satisfies $h_a(\theta)>0$ for all $\theta \in \Theta$. Letting $H_{-a}$ be the marginal distribution of $\thetaa$, that has density $\prod_{i \neq a } h_i(\theta_i)$ with respect to $\lambda^{\otimes K-1}$, one assumes that 
\[\forall a = 1 ,\dots,K, \ \ \ \ \int_{\Theta^{K-1}} h_a(\theta_a^*) dH_{-a}(\thetaa) < \infty.\]
Under the prior distribution $H$, the Bayes risk of any strategy $\cA$ satisfies
\[\liminf_{T \rightarrow \infty} \frac{\cR^H(T,\cA)}{\log^2(T)} \geq  \frac{1}{2}\sum_{a=1}^K\int_{\Theta^{K-1}} h_a(\theta_a^*) dH_{-a}(\thetaa).\]
\end{theorem}

For exponential family bandit models with a product prior, Lai provides the first (asymptotic) prior-dependent Bayes risk upper bounds, when $\Theta$ is compact. Letting $[\mu_0^-,\mu_0^+]=\dot{b}(\Theta)$, he shows in particular that the index policy based on
\begin{equation}I_a(t) = \sup \left\{q \in [\mu_0^-,\mu_0^+] : N_a(t) \overline{d}(\hat{\mu}_a(t),q) \leq \log\left(\frac{T}{N_a(t)}\right)\right\},
\label{index:Lai}
\end{equation}
where $\overline{d}(x,y) = {d}\left(\max(\mu^-_0,\min(\mu^+_0,x)) , y)\right)$, has a Bayes risk that asymptotically matches the lower bound of Theorem~\ref{thm:LRB}. 
This index policy is very similar to kl-UCB-H$^+$ and differs only from the use of a regularized version of the divergence function $d$. 

While a recent line of research on Bayesian randomized algorithms (e.g. Thompson Sampling) has provided Bayes risk upper bounds in quite general settings (\cite{RussoVanRoy13,RussoVanRoy14IDS}), to the best of our knowledge, no upper bound scaling in $\log^2(T)$ has been obtained for exponential family bandit models since the work of Lai. \cite{BubeckLiu13,LiuLi15} give the first prior-dependent upper bounds on the Bayes risk of Thompson Sampling, in a particular case quite different from our setting: a two-armed bandit model in which the means of the arms are known up to a permutation. The joint prior distribution is thus supported on $(\mu_1,\mu_2)$ and $(\mu_2,\mu_1)$. In Section~\ref{subsec:ExpBayesian}, we investigate numerically the optimality  of the Bayesian index policies discussed in the paper with respect to the lower bound of Theorem~\ref{thm:LRB}.

% \paragraph{Matching algorithms} Assume that a strategy satisfies, for all $\thetaB$ and all $a\neq a^*$, 
% \[\bE_{\thetaB}[N_a(T)] \leq (1+\varepsilon) \frac{\log(T(\theta^*-\theta_a))}{K(\theta_a,\theta^*)} + S_{a,\varepsilon,T}(\thetaB),\]
% and that the term $S_{a,\varepsilon,T}$ satisfies, for all compact $B\subset \Theta$,
% \[\int_{\{\thetaB \in B^K : \theta_a^* - \theta_a > \frac{1}{\sqrt{T\log T}}\}} S_{a,\varepsilon,T}(\thetaB)dH(\thetaB) = o(\log^2(T)).\]
% Then one can prove that, for all compact $B\subset \Theta$,  
% \[\limsup_{T\rightarrow \infty} \frac{\int_{B^{K}} R_T(\cA,\thetaB)dH(\thetaB)}{\log^2(T)} \leq \frac{1}{2}\sum_{a=1}^K\int_{\Theta^{K-1}} h_a(\theta_a^*) dH_{-a}(\thetaa).\]
% (which is not completely what we want...) 

\subsection{Approximation of the Finite-Horizon Gittins indices}

As discussed Section~\ref{sec:Gittins}, the FH-Gittins algorithm, that is the index policy associated to  \[J_a(t) = G(\pi_a^t, T-t),\] is conjectured to be a good approximation of the Bayesian optimal policy, yet the above indices remain difficult to compute. Building on approximations of the Finite-Horizon Gittins indices that can be extracted from the literature permits to obtain a related \emph{efficient} index policy.  

Recall from Definition~\ref{def:FHG} that the Finite-Horizon Gittins index takes the form \[G(\pi, r) = \inf \left\{ \lambda \in \R : V_\lambda^*(\pi, r)=0\right\},\] where 
$V^*_\lambda(\pi, r)$ 
corresponds to the optimal value function associated to a calibration game $\cC_\lambda$. 
In the paper \cite{BurnKat03OneArmed}, Burnetas and Katehakis propose tight bounds on the value function $V^*_\lambda(\pi_{a,n,x}, r)$ for exponential family bandits. These bounds permit to derive asymptotic approximations of the FH-Gittins indices, when $r$ is large, and to show that, for large values of the remaining time $T-t$, 
\begin{equation}J_a(t) \simeq \sup \left\{q \in [\mu^-,\mu^+] : N_a(t) \tilde{d}(\hat{\mu}_a(t),q) \leq \log\left(\frac{T-t}{N_a(t)}\right)\right\}.\label{index:Katehakis}\end{equation}
This approximation is valid under the assumption that $\Theta$ is compact: $[\mu^-,\mu^+]=\dot{b}(\Theta)$ and $\tilde{d}$ is another regularization of the divergence function $d$, such that, for any $y$, $\tilde{d}(x,y)=d(x,y)$ for $x> \mu^-$ and for $x\leq\mu^-$, 
\[\tilde{d}(x,y)=d(\mu^-,y) + (\dot{b}^{-1}(y)-\dot{b}^{-1}(\mu^-))(\mu^- - x).\]
In the particular case of Gaussian bandit models, the work of Chang and Lai \cite{ChangLai87} on the approximation of discounted Gittins indices can also be adapted to obtain approximations of the Finite-Horizon Gittins indices, showing the same tendency as in \eqref{index:Katehakis}: compared to the corresponding kl-UCB index, here the $\log t$ is replaced by $\log((T-t)/N_a(t))$. This alternative exploration rate also appears in the non-asymptotic lower bound on the Gaussian Gittins index obtained by \cite{Tor15Gittins}.

These approximations of the Finite-Horizon Gittins indices provide another justification for exploration rates of the form $\log(h(t,T)/N_a(t))$, with some function $h$, which are also used by the kl-UCB-H$^+$ and kl-UCB$^+$ algorithms. These two algorithms can thus be viewed as Bayesian (inspired) index policies. 

\section{Numerical experiments\label{sec:Experiments}}

\subsection{Regret minimization}

\begin{figure}[h]
\centering
\includegraphics[width=0.45\textwidth]{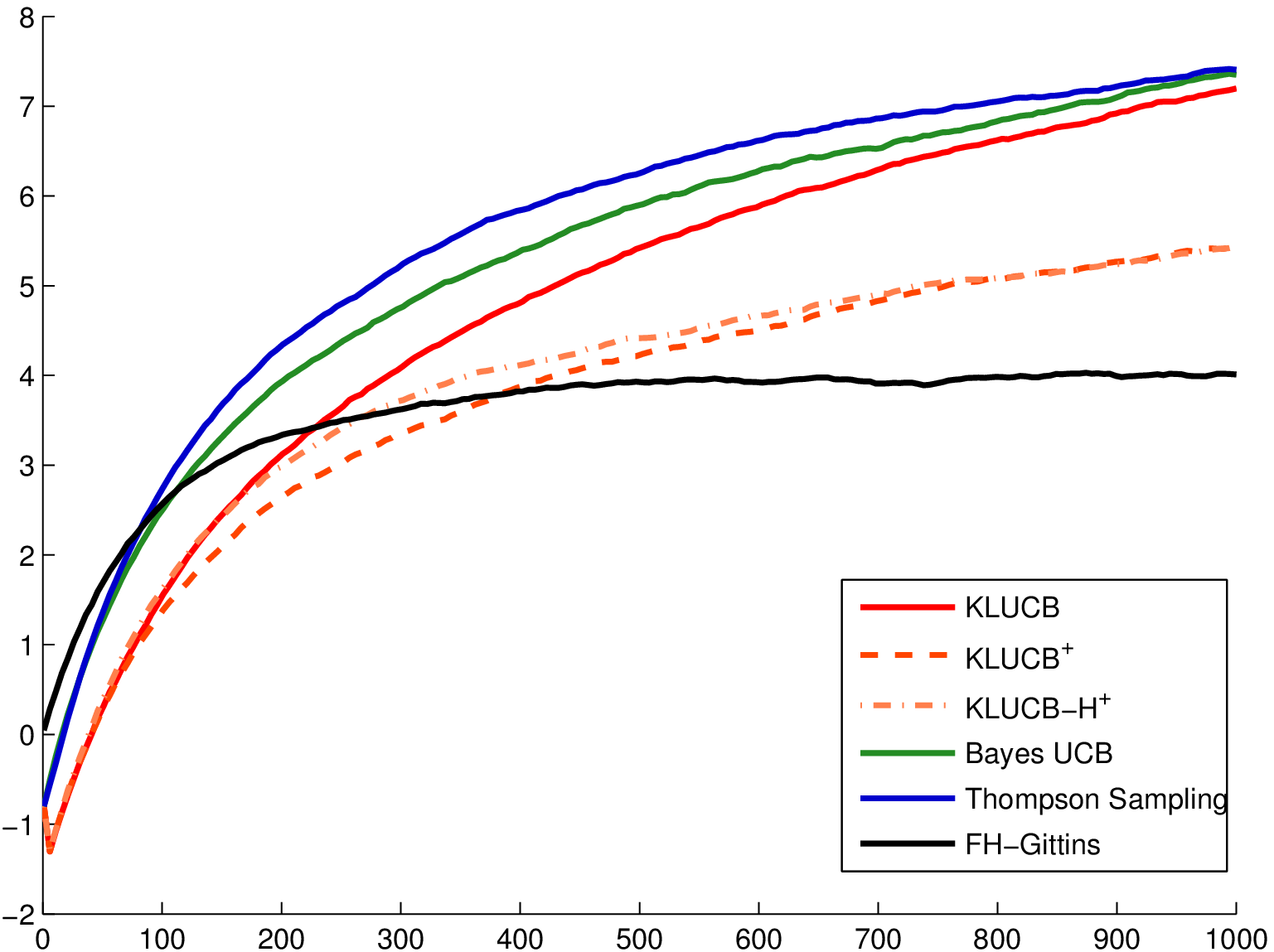}
\includegraphics[width=0.45\textwidth]{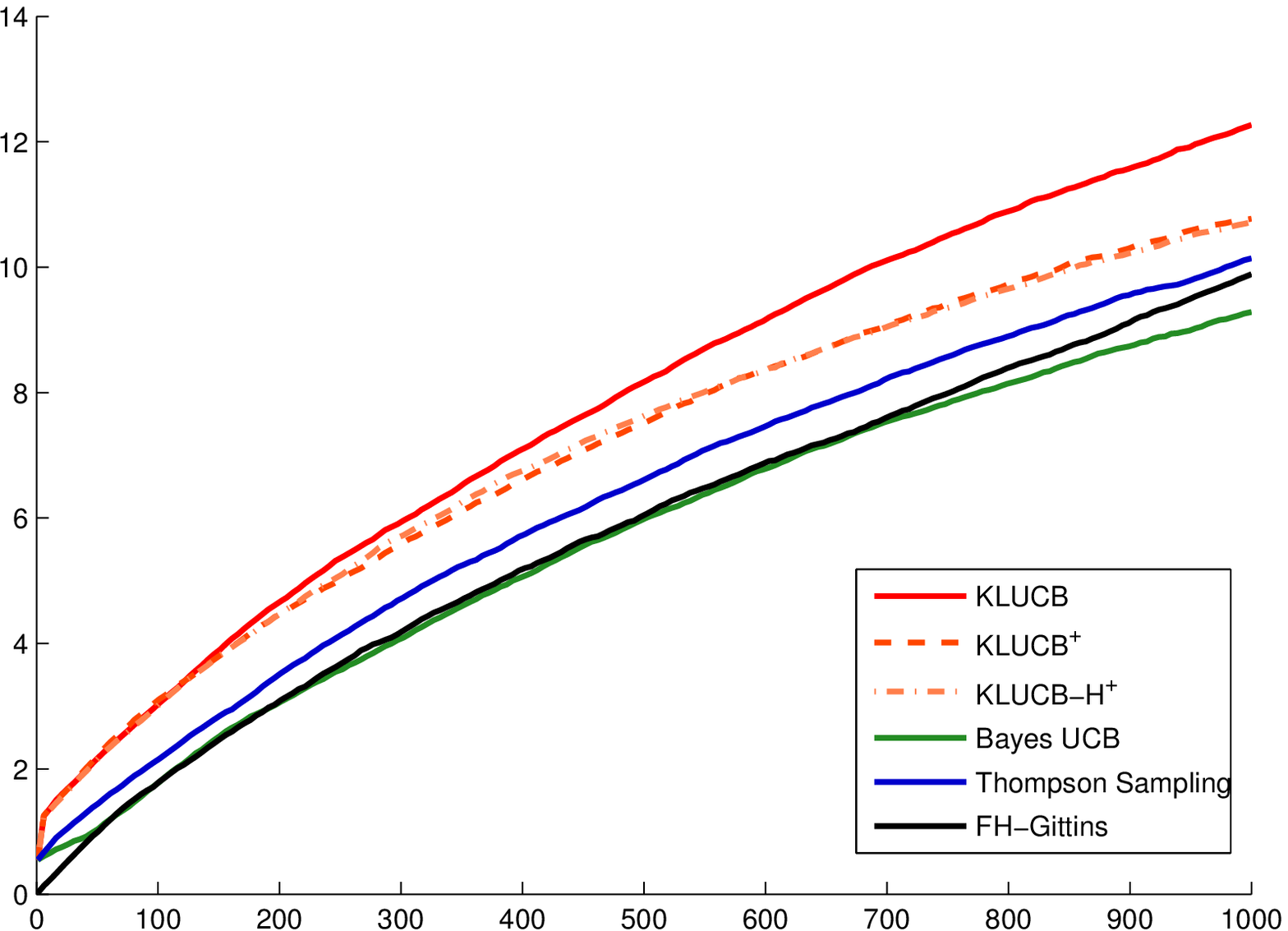} 
\caption{Regret on two-armed Bernoulli bandits ($\bm \mu =[0.05 \ 0.15]$ (left) $\bm \mu = [0.75 \ 0.8]$ (right)) up to horizon $T=1000$, averaged over  $N=10000$ simulations \label{fig:SmallH}}
\end{figure}

We first perform experiments with a moderate horizon $T=1000$, which permits to include the Finite-Horizon Gittins algorithm discussed in Section~\ref{sec:Gittins}. Figure~\ref{fig:SmallH} displays the regret of kl-UCB, Thompson Sampling and the four Bayesian (or Bayesian inspired) index policies discussed in this paper, in two instances of two-armed Bernoulli bandit problems. The Bayesian index policies display comparable, if not better, performance than kl-UCB and Thompson Sampling. In particular, FH-Gittins appears to be significantly better than the other algorithms on the instance with small rewards. 

\begin{figure}[h]
\centering
\includegraphics[width=0.53\textwidth]{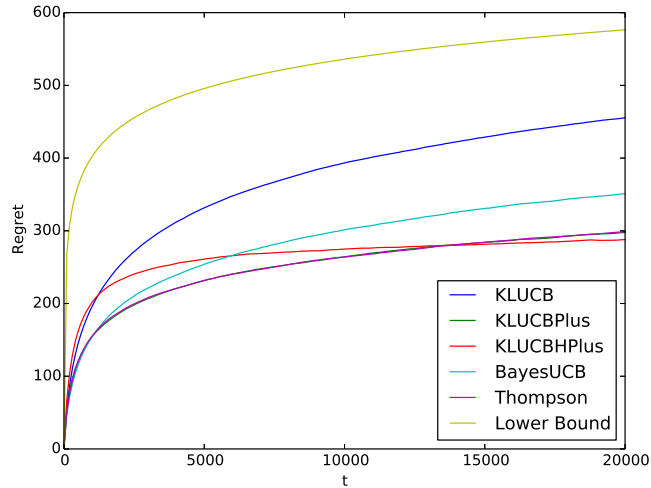}
\caption{Regret on a five-armed bandit with Exponential distributions with means $\bm \mu = [1 \ 1.5 \ 2 \ 2.5 \ 3]$ up to horizon $T=20000$, averaged over $N=50000$ simulations \label{fig:LongH}}
\end{figure}

For a larger horizon $T=20000$, we then run experiments on a bandit model in which rewards follow an exponential distribution (which is a particular Gamma distribution). Bayes-UCB and Thompson Sampling are implemented using a conjugate $\mathrm{InvGamma}(1,1)$ prior on the means. Results are displayed in Figure~\ref{fig:LongH}. In this setting, Bayes-UCB, kl-UCB$^+$ and kl-UCB-H$^+$ improve over kl-UCB, and are also competitive with Thompson Sampling. As already noted in several works (e.g. \cite{KLUCBJournal}), the Lai and Robbins lower bound, that is asymptotic, is quite pessimistic for finite (even large) horizons.

\subsection{Bayes risk minimization\label{subsec:ExpBayesian}}

In this paper, Bayes risk minimization and its exact solution is mostly presented as a justification for improved algorithms for regret minimization. However, it is also  interesting to understand whether the proposed algorithm are good approximations of the Bayesian solution, i.e. whether they match the asymptotic lower bound of Theorem~\ref{thm:LRB}. 

\begin{figure}[h]
\centering
\includegraphics[width=0.45\textwidth]{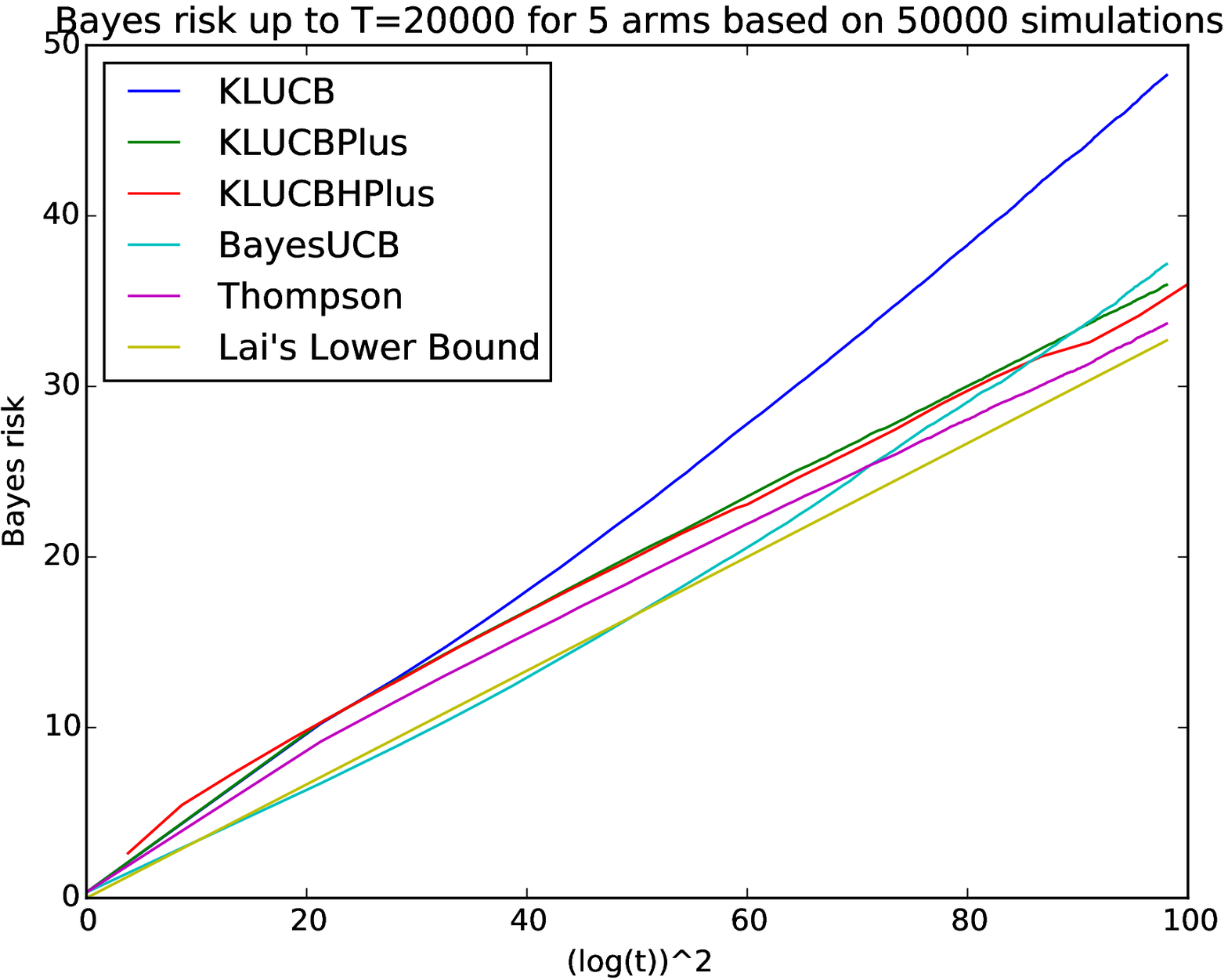}
\includegraphics[width=0.45\textwidth]{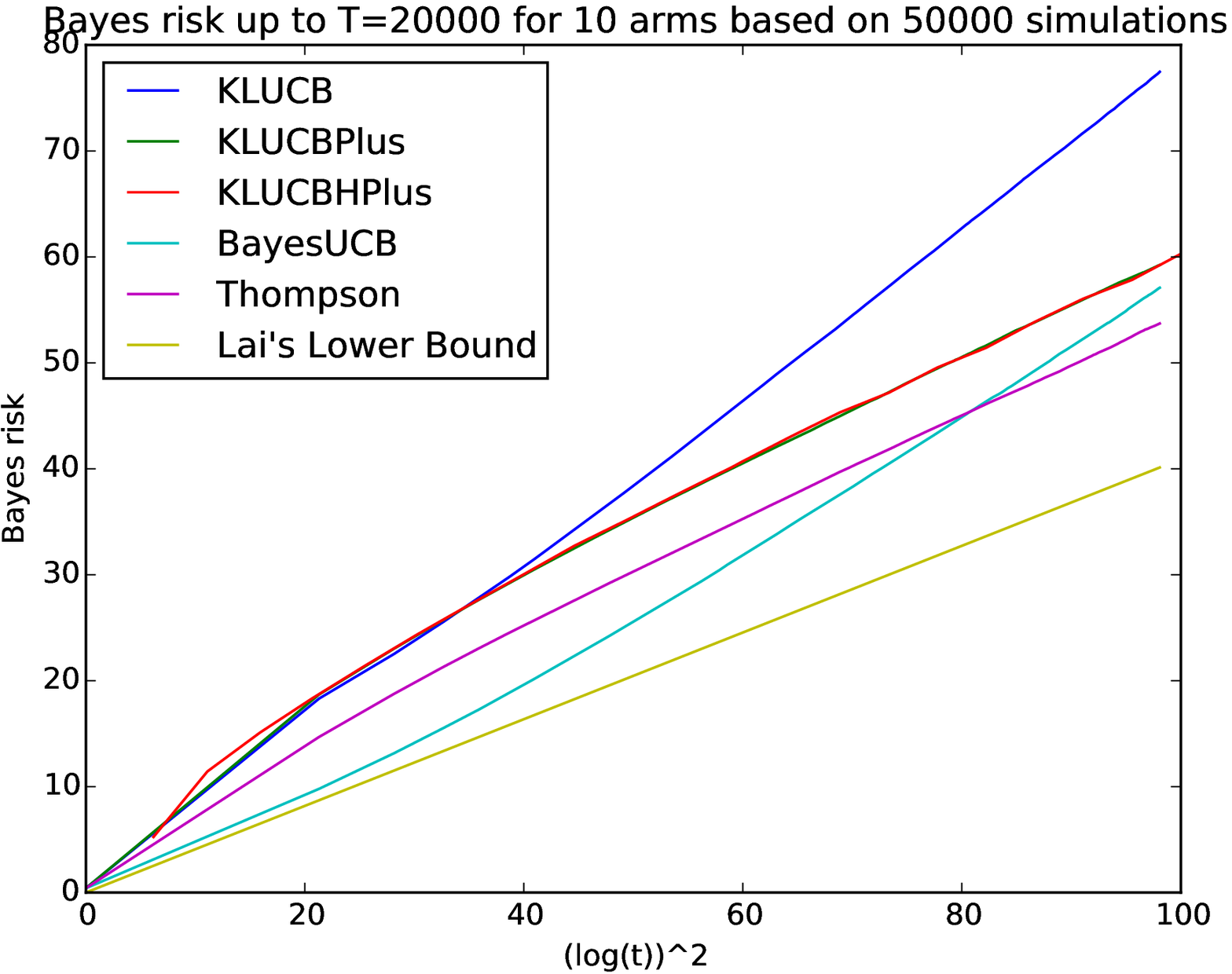} 
\includegraphics[width=0.45\textwidth]{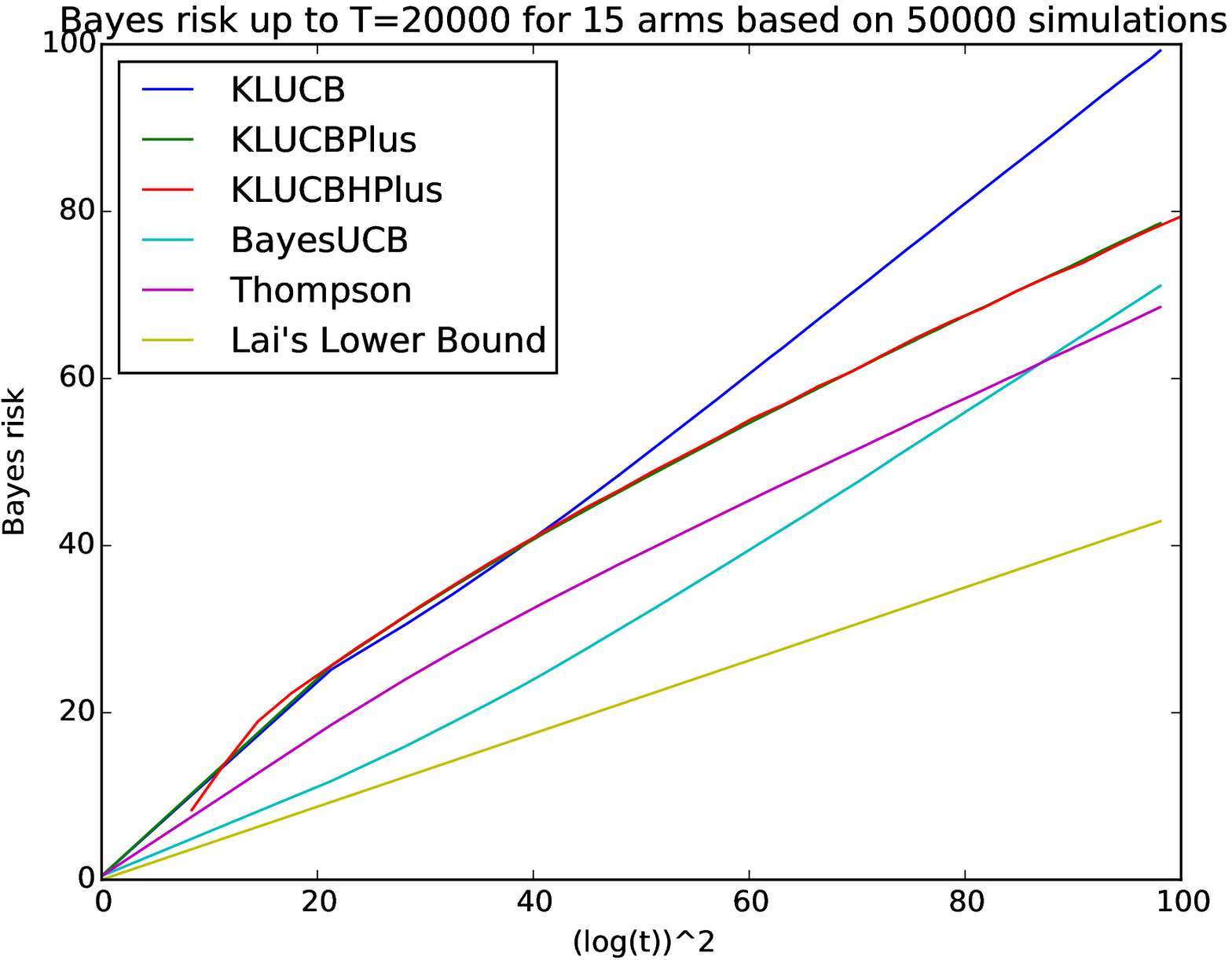} 
\includegraphics[width=0.45\textwidth]{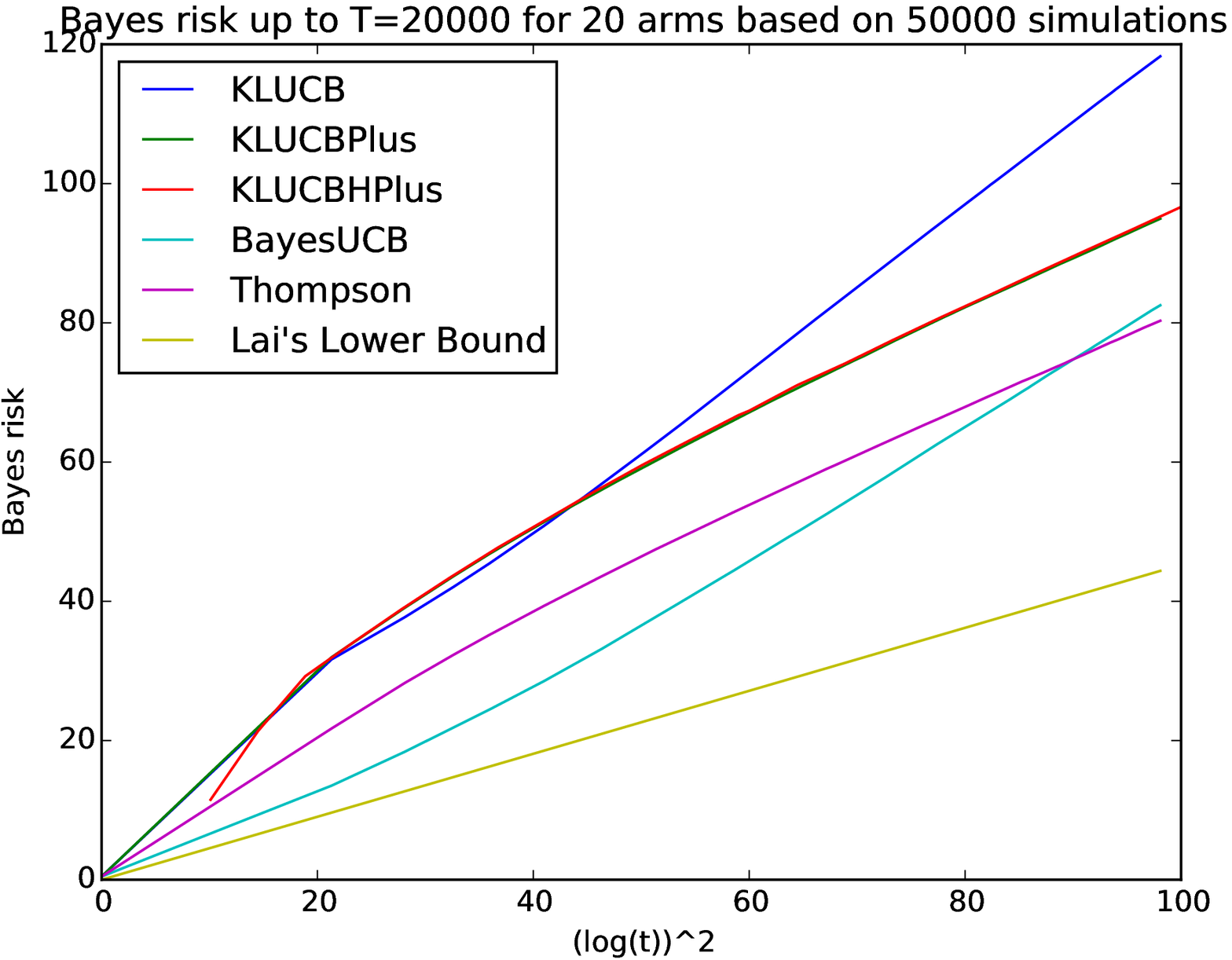}
\caption{Bayes risk up to $T=20000$ on a Bernoulli bandit model with a uniform prior on the $K$ arms, for $K=5,10,15,20$, averaged over $N=50000$ simulations.\label{fig:Bayesian}}
\end{figure}

We report here results of experiments in Bernoulli bandit models with a uniform prior on the means. In this setting, some computations (that are detailled in Appendix~\ref{subsec:BorneBernoulli}) show that the lower bound rewrites 
\[\liminf_{T \rightarrow \infty} \frac{\cR(T,\cA)}{\log^2(T)} \geq \frac{K-1}{K+1}.\]
In particular, we see that the asymptotic rate of the Bayesian regret is (almost) independent of the number of arms. For several values of $K$, we display on Figure~\ref{fig:Bayesian} the Bayes risk $\cR_T(\cA_{(T)})$ of several algorithms, together with the theoretical lower bound,  as a function of $\log^2(T)$. 

For each value of $K$, we observe that all the algorithms have a Bayes risk that seems to be affine in $\log^2(T)$. For Thompson Sampling, kl-UCB$^+$ and kl-UCB-H$^+$ the slope is close to  $(K-1)/(K+1)$, whereas for kl-UCB and Bayes-UCB it is strictly larger. This leads to the conjecture that the first three algorithms are asymptotically optimal in a Bayesian sense. 
It is to be noted that, while the Bayes risk of these algorithms seems to be of order $(K-1)/(K+1)\log^2(T) + C(K)$ for large values of $T$, the second-order term $C(K)$ appears to be increasing significantly with the number of arms. Compared to Lai and Robbins' lower bound on the regret, this lower bound does not appear to be over-pessimistic in finite time.

\section{Conclusion} This paper provides an analysis of the Bayes-UCB algorithm that does not rely on arguments specific to Bernoulli or Gaussian distributions, and is valid in any exponential family bandit model. It also brings theoretical justifications for the use of the kl-UCB-H$^+$ and kl-UCB$^+$ algorithms together with a new insight on the alternative exploration rate used by these algorithms. Finally, the proposed analysis holds for a wide class of prior distributions, namely all distributions that have positive density with respect to the Lebesgue measure. This shows that the choice of prior has no impact on the asymptotic optimality of Bayes-UCB, unlike what happens for Thompson Sampling in Gaussian bandit with unknown mean and variance \cite{HondaTakemura14TS}. Beyond asymptotic optimality, an interesting direction of future work would be to quantify the impact of the prior on second-order terms in the regret. Another important research direction is to better understand the Finite-Horizon Gittins strategy, which performs well in practice, but whose asymptotic optimality is still to be established.

\paragraph{Acknowlegement} The author acknowledges the support of the French Agence Nationale de la Recherche (ANR) under grants ANR-13-BS01-0005 (SPADRO project) and ANR-11-JS02-005-01 (GAP project).

\bibliography{biblioBandits}

\newpage

\hspace{0.3cm}

\appendix 

The appendix is structured as follows. Appendix \ref{section:KL} presents Pinsker-like inequalities, that is quadratic approximations of the Kullback-Leibler divergence functions, when the natural parameters of the distributions belong to some compact interval. These inequalities will be useful at several places of this supplementary material. Appendix~\ref{proof:FiniteTime} gathers the proof of Theorem~\ref{thm:KLvariants} and the proofs of the lemmas introduced in the finite-time analysis of Bayes-UCB. Appendix~\ref{section:TailBounds} and Appendix~\ref{proof:LRB} are respectively dedicated to establishing the posterior tail bounds of Lemma~\ref{lem:DeterministicULB} and proving the asymptotic lower bound on the Bayes risk stated in Theorem~\ref{thm:LRB}.

\section{Pinsker-like inequalities}\label{section:KL}

For on any compact $\cC \subset \Theta$, one can obtain quadratic approximations of the KL-divergence as a function of either the natural parameters or the means. These useful inequalities are stated in Proposition~\ref{prop:Pinsker}

\begin{proposition}\label{prop:Pinsker} Let $\cC$ be a compact subset of $\Theta$. Introducing 
\begin{equation}c_1 := \inf_{\theta \in \cC} \ \ddot{b}(\theta)>0 \ \ \ \ \text{and} \ \ \ \ \ c_2 := \sup_{\theta \in \cC} \ \ddot{b}(\theta)< \infty,\label{def:c1c2}\end{equation}
one has 
\begin{eqnarray}
\forall (\theta,\theta') \ \in (\cC)^2, \ \  \frac{c_1}{2}(\theta-\theta')^2 \leq & K(\theta,\theta') \leq & \frac{c_2}{2}(\theta - \theta')^2, \label{LagrangeK}\\
\forall (x,v) \ \in (\dot{b}(\cC))^2, \ \  \frac{1}{2c_2} (x-v)^2 \leq &d(x,v) \leq& \frac{1}{2c_1}(x-v)^2. \label{LagrangeD}
\end{eqnarray}
If $(x,v)\in (\dot{b}(\cC))^2$ are such that $x<v$, one has  
\begin{equation}{\dot b}^{-1}(v)-{\dot b}^{-1}(x)\leq \frac{1}{c_1} (v-x).\label{PinskerL}\end{equation}
\end{proposition}

\paragraph{Proof}
These three statements follow from Lagrange formulas. For example to derive \eqref{LagrangeD}, given that $d(x,y)=\mathrm{K}({\dot b}^{-1}(x),\dot{b}^{-1}(y))$, it can be shown, using the close form expression \eqref{KLClose}, that 
\[\frac{d}{dx}d(x,v) = \dot{b}^{-1}(x) - \dot{b}^{-1}(v) \ \ \ \text{and} \ \ \ \frac{d^2}{d^2x}d(x,v) = \frac{1}{\ddot{b}(\dot{b}^{-1}(x))}.\]
From the second-order Lagrange formula applied to $x \mapsto d(x,v)$, there exists $c \in ]x,v[$ (or $]v,x[$) such that 
\[d(x,v) = \frac{1}{2}\frac{1}{{\ddot b}({\dot b}^{-1}(c))} (x-v)^2 \leq \frac{1}{2c_1}(x-v)^2.\]
The other inequalities are obtained using similar arguments.

\section{Finite-time analysis} \label{proof:FiniteTime}

\subsection{Proof of Theorem~\ref{thm:KLvariants}} 

We first give an analysis of the index policy associated to $u_a^+(t)$. Introducing $g_t$ defined by  $d(\mu_1-g_t,\mu_1)=\frac{1}{\log(t)}$, one can write a decomposition similar to that used in the proof of Theorem~\ref{thm:BayesUCB}: 
\begin{eqnarray}
\bE[N_a(T)] &&\leq 1 + \sum_{t=K}^{T-1}\bP\left(\mu_1-g_t \geq u_1^+(t)\right) \nonumber\\
&&+ \sum_{t=K}^{T-1}\bP(\mu_1 - g_t \leq u_a^+(t), A_{t+1}=a) \nonumber\\
 & & \leq  1 +  \sum_{t=K}^{T-1}\bP\left(N_1(t)d^+(\hat{\mu}_1(t), \mu_1-g_t) \geq \log \left(\frac{t\log^c(t)}{N_a(t)}\right)\right) \label{step:sum1}\\
  & &+ \sum_{t=K}^{T-1}\bP\left(N_a(t)d^+(\hat{\mu}_a(t), \mu_1-g_t) \leq \log \left({T\log^c(T)}\right),A_{t+1}=a\right),\label{step:sum2}
\end{eqnarray}
using the definition of $u_a^+(t)$ and the fact that $t\log^c t/N_a(t) \leq T \log^c T$.
Lemma~\ref{lem:TermA} can be applied (with $A=1$) to show that the sum in \eqref{step:sum1} is of order $o(\log(T))$, while the sum in \eqref{step:sum2} can be rewritten and upper bounded using Lemma~\ref{lem:TermB}: for all $\varepsilon>0$, the result follows from 
\begin{align*}
& \bE \sum_{t=K}^{T-1}\sum_{s=1}^t\ind_{\left(sd^+(\hat{\mu}_{a,s}, \mu_1 - g_t ) \leq \log(T \log^c T)\right)} \ind_{(A_{t+1}=a, N_a(t) = s)} \\
 & \leq \sum_{s=1}^{T-1}\bP\left(sd^+(\hat{\mu}_{a,s}, \mu_1-g_t) \leq \log \left({T\log^c(T)}\right)\right)\\
& \leq  \frac{1+\varepsilon}{d(\mu_a,\mu_1)}\log(T\log^c(T)) + o_{\varepsilon}(\log(T)).
\end{align*}

For the index policy associated to $u_a^{H,+}(t)$, using a similar decomposition,
\begin{eqnarray}
\bE[N_a(T)] && \leq  1 + \sum_{t=K}^{T-1}\bP\left(\mu_1-g_t \geq u_1^{H,+}(t)\right) \nonumber \\
&&+ \sum_{t=K}^{T-1}\bP(\mu_1 - g_t \leq u_a^{H,+}(t), A_{t+1}=a) \nonumber\\
 && \leq  1 +  \sum_{t=K}^{T-1}\bP\left(N_1(t)d^+(\hat{\mu}_1(t), \mu_1-g_t) \geq \log \left(\frac{T\log^c(T)}{N_a(t)}\right)\right) \label{step:sum3}\\
 & & \ \ + \sum_{t=K}^{T-1}\bP\left(N_a(t)d^+(\hat{\mu}_a(t), \mu_1-g_t) \leq \log \left({T\log^c(T)}\right),A_{t+1}=a\right)\label{step:sum4}.
\end{eqnarray}
The sums in \eqref{step:sum4} and \eqref{step:sum2} are the same, and lower bounding $T\log^c T $ by $t\log^c t$ in each term of the sum in \eqref{step:sum3} shows that it is upper bounded by \eqref{step:sum1}. Thus, this index policy is also asymptotically optimal.

\subsection{Proof of Lemma~\ref{lem:TermA}} To upper bound 
\[(A):=\bP\left(N_1(t) d^+(\hat{\mu}_1(t),\mu_1-g_t) \geq \log \frac{At\log^c t}{N_1(t)}\right),\]
we consider two cases in which arm 1 has or not been drawn a lot.  
\begin{align*}
  (A)  &\leq  \underbrace{\bP\left(N_1(t) d^+(\hat{\mu}_1(t),\mu_1-g_t) \geq \log\left(\frac{At\log^c t}{N_1(t)}\right), N_1(t) \leq \log^4(t)\right)}_{A_1} \\
  &  + \underbrace{\bP\left(N_1(t) d^+(\hat{\mu}_1(t),\mu_1-g_t) \geq \log\left(\frac{At\log^c t}{N_1(t)}\right), N_1(t) > \log^4(t)\right)}_{A_2}
\end{align*}
To upper bound term $A_1$, we write 
\begin{eqnarray*}
& & \left(N_1(t) d^+(\hat{\mu}_1(t),\mu_1-g_t) \geq \log\left(\frac{At\log^c t}{N_1(t)}\right),  N_1(t) \leq \log^4(t)\right)\\
& & \subseteq  \left(N_1(t)d^+\left(\hat{\mu}_1(t),\mu_1\right)\geq  \log(At) + c\log\log(t) -4\log \log(t) ,  \right. \\
& &\hspace{8cm} N_1(t) \leq \log^4(t)\Big)\\
& & \subseteq \left(N_1(t)d^+\left(\hat{\mu}_1(t),\mu_1 \right)\geq \log(At) + 3\log \log t\right),
\end{eqnarray*}
using that $c\geq 7$. The self-normalized concentration inequality proved in \cite{KLUCBJournal} and stated in Lemma \ref{lem:SelfNormAdapt} permits to further upper bound $A_1$: 
\[(A_1)  \leq  e \frac{\log^2 t + 3(\log t) \log \log (t) + \log(A)\log t  + 1}{At \log^3 t}.\]

\begin{lemma}\label{lem:SelfNormAdapt}
\[\bP\left(\exists s \in\{1,\dots,t\}: sd^+\left(\mu_{1,s},\mu_1\right) \geq \delta\right) \leq \left(\delta
    \log(t) + 1\right)\exp(-\delta + 1).\]
\end{lemma}

To upper bound term $A_2$, if $t$ is such that $\log^7 t \geq A^{-1}$, we write 
\begin{eqnarray*}
& & \left(N_1(t) d^+(\hat{\mu}_1(t),\mu_1-g_t) \geq \log\left(\frac{At\log^c t}{N_1(t)}\right), N_1(t) \geq \log^4(t)\right)\\
& & \subseteq  \left(N_1(t)d^+\left(\hat{\mu}_1(t),\mu_1-g_t\right)\geq 0 , N_1(t) \geq \log^4(t)\right) \\
& & \subseteq \left(\hat{\mu}_1(t) \leq \mu_1-g_t , N_1(t) \geq \log^4(t) \right),
\end{eqnarray*}
Thus, if $t$ is such that $t \geq \exp(\sqrt{3})$ (which implies $\log^3 t \geq 3\log t$),
\begin{eqnarray*}
(A_2) & \leq & \bP\left(\hat{\mu}_1(t) \leq \mu_1-g_t , N_1(t) \geq \log^4(t)\right)  \\ 
& \leq & \bP\left(\exists s \in [\lceil\log(t)^4\rceil; t] \ : \ \hat{\mu}_{1,s} \leq \mu_1 - g_t\right) \\
& \leq &\sum_{s=\lceil\log(t)^4\rceil}^t\bP\left(\hat{\mu}_{1,s} \leq \mu_1 - g_t\right) \leq \sum_{s=\lceil\log(t)^4\rceil}^t e^{-sd(\mu_1 - g_t,\mu_1)}\\
& \leq & t e^{-(\log t)^4 d(\mu_1 - g_t,\mu_1)} = t e^{-(\log t)^3} \leq t e^{-3 \log t} = \frac{1}{t^2}.
\end{eqnarray*}
Combining this with the upper bound on $A_1$ yields the result.

\subsection{Proof of Lemma~\ref{lem:TermB}}

The quantity to be upper bounded is
\[
(B) := \sum_{s=1}^{T} \bP\left(sd^+\left(\hat{\mu}_{a,s},\mu_1-g(s)\right) \leq f(T) + h(s)\right). 
\]
The function $w(q)=d^+(\hat{\mu}_{a,s},q)$ is convex.  Moreover it is differentiable on the interval $]\hat{\mu}_{a,s},\mu^+[$ with
$w'(q)=\frac{q-\hat{\mu}_{a,s}}{\mathrm{V}(q)}\ind_{(\hat{\mu}_{a,s}\leq q)}$. Thus, if $\hat{\mu}_{a,s} \geq \mu_0^-$, 
$$  d^+(\hat{\mu}_{a,s},\mu_1 - g(s)) \geq  d^+(\hat{\mu}_{a,s},\mu_1)  - g(s)\frac{\mu_1-\hat{\mu}_{a,s}}{\mathrm{V}(\mu_1)} \geq 
d^+(\hat{\mu}_{a,s},\mu_1)  - g(s)\frac{\mu_1 - \mu_0^-}{\mathrm{V}(\mu_1)}\;.$$
Therefore
\begin{eqnarray*}
(B)&\leq&\sum_{s=1}^{T}\bP\left(d^+(\hat{\mu}_{a,s},\mu_1) \leq \frac{f(T)}{s} + g(s)\frac{\mu_1-\mu_0^-}{\mathrm{V}(\mu_1)} + \frac{h(s)}{s}\right) + \sum_{s=1}^{T}\bP(\hat{\mu}_{a,s} < \mu_0^-)\\
& \leq & \sum_{s=1}^{T}\bP\left(d^+(\hat{\mu}_{a,s},\mu_1) \leq \frac{f(T)}{s} + r(s)\right) + \frac{1}{1-e^{-d(\mu_0^-,\mu_a)}},
\end{eqnarray*}
using Chernoff inequality and introducing  
\[r(s):=g(s)\frac{\mu_1-\mu_0^-}{\mathrm{V}(\mu_1)} + \frac{h(s)}{s}.\]
Let $\varepsilon>0$. One also introduce
\[K_T(\varepsilon) := \left\lceil \frac{(1+\varepsilon)f(T)}{d(\mu_a,\mu_1)} \right\rceil.\]
From the assumptions on $f,g$ and $h$, there exists $s_0$ such that $r$ is non-increasing for $s\geq s_0$ and one has 
\[K_T(\varepsilon) \underset{T \rightarrow \infty}{\longrightarrow} \infty \ \ \ \text{and} \ \ \ r(s) \underset{s \rightarrow \infty}{\longrightarrow} 0.\]
For $T$ such that $K_T \geq s_0$,
\[(B) \leq K_T + \sum_{s=K_T +1}^{T}\bP\left(d^+(\hat{\mu}_{a,s},\mu_1) \leq \frac{f(T)}{s} + r(K_T)\right) + C_a,\]
with $C_a=1/\left(1-e^{-d(\mu_0^-,\mu_a)}\right)$. As $r(K_T) \rightarrow 0$, there exists $N_a(\varepsilon)$ such that 
\[T \geq N_a(\varepsilon) \ \Rightarrow \ r(K_T) \leq d(\mu_a,\mu_1)\frac{\varepsilon}{1+\varepsilon}.\]
Then, if $T\geq N_a(\varepsilon)$, one has, for all  $s \geq K_T +1$, 
\[\frac{f(T)}{s} + r(K_T) \leq d(\mu_a,\mu_1)\]
and there exists $\mu^*(s)\in]\mu_a;\mu_1[$ 
such that $d(\mu^*(s),\mu_1)=\frac{f(T)}{s} +   r(K_T)$.
Then, using Chernoff inequality and the inequality 
\[\forall \mu > \mu', \ \ d(\mu,\mu') \geq \frac{1}{2 \sup_{\mu \in [\mu',\mu]}\mathrm{V}(\mu)} (\mu - \mu')^2,\]
stated in \cite{KLUCBJournal} and that follows from Lagrange equality, one can write
\begin{eqnarray*}
 (B) & \leq & K_T + \!\!\!\sum_{s=K_T+1}^T \!\!\bP(\hat{\mu}_{a,s} > \mu^*(s)) + C_a \leq K_T + \!\!\!\sum_{s=K_T+1}^T e^{-sd(\mu^*(s),\mu_a)} + C_a\\
     & \leq & K_T + \!\!\!\sum_{s=K_T+1}^T \! e^{-s\frac{(\mu^*(s)-\mu_a)^2}{2\mathrm{V}_a^2}} + C_a \leq K_T + \int_{K_T}^{\infty} e^{-s\frac{(\mu^*(s)-\mu_a)^2}{2\mathrm{V}_a^2}}ds + C_a, 
\end{eqnarray*}
where $\mathrm{V}_a = \sup_{\mu \in ]\mu_a,\mu_1[} \mathrm{V}(\mu)$. 
Using the convexity of $x \mapsto d(x,\mu_1)$, a lower bound on $\mu^*(s)-\mu_a$ 
can be obtained, as in Appendix 2 of \cite{KLUCBJournal}: 
\[\mu^*(s)-\mu_a \geq \frac{d(\mu_a,\mu_1) - \left[\frac{f(T)}{s}+r(K_T)\right]}{-d'(\mu_a,\mu_1)}\]
 \cite{KLUCBJournal} also provide tight upper bound on the resulting integrals, and following a similar approach
 allows us to conclude the proof:
\begin{align*}
&\int_{K_T}^{\infty} e^{-s\frac{(\mu^*(s)-\mu_a)^2}{2\mathrm{V}_a^2}}ds \\
& \leq  \int_{K_T}^{\infty}\exp\left(- \frac{s}{2\mathrm{V}_a^2d'(\mu_a,\mu_1)^2}\left(d(\mu_a,\mu_1)- \left(\frac{f(T)}{s}+r(K_T)\right)\right)^2\right)ds \\
& \leq  f(T)\int_{\frac{1+\varepsilon}{d(\mu_a,\mu_1)}}^\infty 
\exp\left(- \frac{uf(T)}{2\mathrm{V}_a^2d'(\mu_a,\mu_1)^2}\left(d(\mu_a,\mu_1)- \left(\frac{1}{u}+r(K_T)\right)\right)^2\right)du \\
& \leq  f(T)\int_{\frac{1+\varepsilon}{d(\mu_a,\mu_1)}}^{\frac{2(1+\varepsilon)}{d(\mu_a,\mu_1)}} 
\exp\left(- \frac{(1+\varepsilon)\left(d(\mu_a,\mu_1)-\left(\frac{1}{u}+r(K_T)\right)\right)^2}{2\mathrm{V}_a^2d(\mu_a,\mu_1)d'(\mu_a,\mu_1)^2}f(T)\right)du \\
&  \hspace{0.5cm} + f(T)\int_{\frac{2(1+\varepsilon)}{d(\mu_a,\mu_1)}}^\infty 
\exp\left(- \frac{uf(T)}{2\mathrm{V}_a^2d'(\mu_a,\mu_1)^2}\frac{d(\mu_a,\mu_1)^2}{4(1+\varepsilon)^2}\right)du \\
& \leq  f(T)\frac{4(1+\varepsilon)^2}{d(\mu_a,\mu_1)^2}\int_{0}^{\infty} 
\exp\left(- \frac{(1+\varepsilon)v^2f(T)}{2\mathrm{V}_a^2d(\mu_a,\mu_1)d'(\mu_a,\mu_1)^2} \right)dv \\
& \hspace{0.5cm}+ 8(1+\varepsilon)^2\mathrm{V}_a^2 \left(\frac{d'(\mu_a,\mu_1)}{d(\mu_a,\mu_1)}\right)^2 \\
& \leq  \sqrt{f(T)} \sqrt{\frac{8\mathrm{V}_a^2\pi(1+\varepsilon)^3d'(\mu_a,\mu_1)^2}{d(\mu_a,\mu_1)^3}} 
+ 8(1+\varepsilon)^2 \mathrm{V}_a^2\left(\frac{d'(\mu_a,\mu_1)}{d(\mu_a,\mu_1)}\right)^2.
\end{align*}

\section{Posterior tail bounds} \label{section:TailBounds}

Let $\mu^-, \mu^+$ be such that $\JJ := \dot{b}(\Theta) = (\mu^-,\mu^+)$. We give here the proof of Lemma~\ref{lem:DeterministicULB}, that follows directly from bounds on 
\[\pi_{n,x}([v, \mu^+[) := \frac{\int_{v}^{\mu^+}\exp(-nd(x,u)) f_0(u)du}{\int_{\JJ}\exp(-nd(x,u)) f_0(u)du}\]
for a density function $f_0$ satisfying $f_0(u)>0$ for all $u\in \JJ$. We fix $\mu_0^-,\mu_0^+$: $\dot{b}(\theta^-) < \mu_0^- < \mu_0^+ < \dot{b}(\theta^+)$.

First, we fix a compact $\cC$ included in $\Theta$ such that $[\mu_1^-,\mu_1^+] := \dot{b}(\cC)$ satisfy 
\[\dot{b}(\theta^-) < \mu_1^- < \mu_0^- < \mu_0^+ < \mu_1^+ <\dot{b}(\theta^+).\]
We let $\mathrm{J}_{\cC} = [\mu_1^-,\mu_1^+]$ and $c_1$ and $c_2$ be the upper and lower bounds on $\ddot{b}$ on $\cC$, defined as \eqref{def:c1c2}. We will often use the quadratic bounds on the Kullback-Leibler divergence on this compact, that are stated in Proposition~\ref{prop:Pinsker}. Also, we will use monotonicity properties of the divergence function: for all $y\in\JJ$, $x \mapsto d(x,y)$ is decreasing on $]\mu^-,y[$ and increasing on $]y,\mu^+[$ whereas for all $x\in \JJ$, $y \mapsto d(x,y)$ is decreasing on $]\mu^-,x[$ and increasing on $]x,\mu^+[$.

Let $x,v$ such that $\mu_0^-<x<v<\mu_0^+$. One has  
\begin{equation}
\pi_{n,x}([v,\mu^+[)  =  \frac{\int_{v}^{\mu^+}e^{-nd(x,u)}f_0(u)du}{\int_{\mu^-}^{\mu^+}e^{-nd(x,u)}f_0(u)du}. \label{Posterior}
\end{equation}
For any $V_{n,x} \subset \mathrm{J}$,
\begin{eqnarray*}
\pi_{n,x}([v,\mu^+[)  & \leq & \frac{e^{-nd(x,v)}\int_{v}^{\mu^+}f_0(u)du}{\int_{V_{n,x}}e^{-nd(x,u)}f_0(u)du} \leq  \frac{e^{-nd(x,v)}}{\int_{V_{n,x}}e^{-nd(x,u)}f_0(u)du}.
\end{eqnarray*}
We now choose $V_{n,x}=\{ u\in \Jc : \frac{n(x-u)^2}{2c_1} \leq 1\}$. From \eqref{LagrangeD}, $nd(x,u)\leq 1$ on $V_{n,x}$. Hence 
\begin{eqnarray*}
\int_{V_{n,x}}e^{-nd(x,u)}f_0(u)du & \geq & e^{-1} \inf_{u \in \Jc}f_0(u) \int_{V_{n,x}}1du
\end{eqnarray*}
and 
\begin{eqnarray*}
\int_{V_{n,x}}1du & = & \lambda\left([\mu_1^-,\mu_1^+]\cap\left[x-\sqrt{\frac{2c_1}{n}},x+\sqrt{\frac{2c_1}{n}}\right]\right) \\ 
& \geq & \min\left(\sqrt{\frac{2c_1}{n}},\mu_1^+-\mu_1^-\right). 
\end{eqnarray*}
The following inequality yields the upper bound in statement 1:
\begin{eqnarray*}
 \pi_{n,x}([v,\mu^+[) & \leq & \frac{e}{\sqrt{2c_1}\inf_{u \in \Jc}f_0(u)}\max\left(\sqrt{n},\frac{\sqrt{2c_1}}{(\mu_1^+-\mu_1^-)}\right)e^{-nd(x,v)}.
\end{eqnarray*}

As  $e^{-nd(x,u)} \leq 1$, the denominator in \eqref{Posterior} is upper bounded by 1, thus 
\[
 \pi_{n,x}([v,\mu^+[) \geq \int_{v}^{\mu^+}e^{-nd(x,u)}f_0(u)du \geq \int_{v}^{\mu_1^+}e^{-nd(x,u)}f_0(u)du.
\]
This last integral can be lower bounded in the following way:
\begin{eqnarray}
\int_{v}^{\mu_1^+}e^{-nd(x,u)}f_0(u)du & = &  e^{-nd(x,v)}\int_{v}^{\mu_1^+}e^{-n\left[d(x,u) - d(x,v)\right]}f_0(u)du \nonumber \\
& = & e^{-nd(x,v)}\int_{v}^{\mu_1^+}e^{-n\left[d(v,u) +(\dot{b}^{-1}(u) -\dot{b}^{-1}(v))(v-x)\right]}f_0(u)du \nonumber\\
& \geq & e^{-nd(x,v)}\int_{v}^{\mu_1^+}e^{-n\left[\frac{1}{2c_1}(u-v)^2 + \frac{1}{c_1} (u -v)(v-x)\right]}f_0(u)du, \label{TBC1}
\end{eqnarray}
where the last inequality follows from \eqref{LagrangeD} and \eqref{PinskerL}. We let 
\begin{eqnarray*}
\phi(u)&=&\frac{1}{2c_1}\left[(u-v)^2 + 2 (u -v)(v-x)\right]. 
\end{eqnarray*}
One has $\phi'(u)=(u-x)/c_1$, thus $\phi$ is strictly increasing on $[v,\mu_1^+]$ and it can be checked that 
$\phi^{-1}(y) = x + \sqrt{(v-x)^2 + 2c_1{y}}$. Thus letting $y=n\phi(u)$, one has 
\[du = \frac{c_1}{n\sqrt{(v-x)^2 + {2c_1y}/{n}}}dy,\]
and 
\begin{align*}
&\eqref{TBC1} = \frac{c_1e^{-nd(x,v)}}{n}\int_{0}^{\frac{n}{2c_1}\left[(\mu_1^+-v)^2 + 2(\mu_1^+-v)(v-x)\right]}\frac{e^{-y}}{\sqrt{(v-x)^2+{2c_1 y}/n}}f_0(y)dy \\ 
&\geq \frac{c_1e^{-nd(x,v)}\min_{\Jc}f_0}{n}\int_{0}^{\frac{n}{2c_1}\left[(\mu_1^+-v)^2 + 2(\mu_1^+-v)(v-x)\right]}\hspace{-3cm}\frac{e^{-y}dy}{\sqrt{(v-x)^2+(\mu_1^+-v)^2 + 2(\mu_1^+-v)(v-x) }} \\ 
& \geq  \frac{c_1e^{-nd(x,v)}\min_{\Jc}f_0}{n(\mu_1^+-x)}\left(1 - e^{-\frac{n}{2c_1}(\mu_1^+-v)^2}\right). 
\end{align*}
Finally, using that $\mu_0^-<x$ and $v \leq \mu_0^+$, one obtains 
\[\pi_{n,x}([v,\mu^+[) \geq \left(\frac{c_1(1 - e^{-\frac{(\mu_1^+-\mu_0^+)^2}{2c_1}})\min_{\Jc}f_0}{(\mu_1^+-\mu_0^-)}\right) \frac{1}{n}e^{-nd(x,v)},\]
which yields the lower bound in statement 1.

We now prove statement 2. Let $x,v$ such that $\mu_0^-<v\leq x < \mu_0^+$. As $[x,\mu_1^+]\subset [v,\mu^+[$, one has 
\begin{eqnarray*}
 \pi_{n,x}([v,\mu^+[) &\geq & \int_{x}^{\mu_1^+} e^{-nd(x,u)}f_0(u)du
\end{eqnarray*}
$F_x : u \mapsto \sqrt{d(x,u)}$ is a one-to-one mapping between $[x,\mu_1^+]$ and $[0,\sqrt{d(x,\mu_1^+)}]$. Moreover, letting $d'(x,u)=\frac{d}{du}d(x,u) = \frac{u-x}{\ddot{b}(b^{-1}(u))} = \frac{u-x}{\mathrm{V}(u)}$, 
\[F'_x(u) = \frac{d'(x,u)}{2\sqrt{d(x,u)}} \underset{u \rightarrow x}{\sim}\frac{(u-x)/\mathrm{V}(u)}{2\sqrt{\frac{1}{2}(x-u)^2/\mathrm{V}(x)}} \underset{u\rightarrow x}{\longrightarrow}\sqrt{\frac{\mathrm{V}(x)}{2}}.\]
$F'_x$ is continuous on $[x,\mu_1^+]$ and strictly positive, thus the inverse mapping $\phi_x : [0,\sqrt{d(x,\mu_1^+)}] \rightarrow [x,\mu_1^+]$ is well defined and differentiable. 
Letting $u=\phi_x(y)$, one has 
\begin{align*}
&\int_{x}^{\mu_1^+}e^{-nd(x,u)}f_0(u) du =\int_{0}^{\sqrt{d(x,\mu_1^+)}} e^{-ny^2}f_0(\phi_x(y)) \frac{2\sqrt{d(x,\phi_x(y))}}{d'(x,\phi_x(y))}dy \\
& \hspace{2.5cm}\geq  \inf_{u\in\mathrm{J}_C}f_0(u) \frac{2}{\sqrt{n}}\int_{0}^{\sqrt{nd(x,\mu_1^+)}} e^{-y^2} \frac{\sqrt{d(x,\phi_x\left(\frac{y}{\sqrt{n}}\right))}}{d'(x,\phi_x\left(\frac{y}{\sqrt{n}}\right))}dy.
\end{align*}
The mapping $(x,u) \mapsto \sqrt{d(x,u)}/d'(x,u)$ is continuous and strictly positive on the compact set $\cS = \{ (x,u) \in [\mu_0^-,\mu_0^+]\times [\mu_1^-,\mu_1^+] : x \leq u \leq \mu_1^+\}$ therefore, one can define
\[c = \inf_{(x,u) \in \cS} \frac{\sqrt{d(x,u)}}{d'(x,u)}>0.\]
For $n\geq 1$, one has 
\[\int_{x}^{\mu_1^+}e^{-nd(x,u)}f_0(u) du \geq \frac{1}{\sqrt{n}}\left(2c\inf_{u\in\mathrm{J}_C}f_0(u) \int_{0}^{\sqrt{d(\mu_0^+,\mu_1^+)}}e^{-y^2}dy \right).\]
Thus there exists a constant $C=C(\mu_1^-,\mu_1^+,f_0)>0$ such that  
\begin{eqnarray*}
  \pi_{n,x}([v,\mu^+[) & \geq & \frac{C}{\sqrt{n}},
\end{eqnarray*}
which concludes the proof.

\section{Lower bound on the Bayesian regret} \label{proof:LRB}

\subsection{Proof of Theorem~\ref{thm:LRB}}

Let $\cA$ be a bandit algorithm. 
Introducing  \[C_{\text{opt}} = \frac{1}{2}\sum_{a=1}^K\int_{\Theta^{K-1}} h_a(\theta_a^*) dH_{-a}(\thetaa),\] 
we assume that $\cA$ satisfies the following: there exists constants $C>C_{\text{opt}}$ and $T_0>0$ such that  
\begin{equation}
\forall T \geq T_0, \ \cR^H(T, \cA) \leq C (\log T)^2.
 \label{ass:Restrict}
\end{equation}
Note that if $\cA$ does not satisfy the above assumption, the desired conclusion follows directly:
\[\liminf_{T\rightarrow \infty} \frac{\cR^H(T, \cA)}{\log^2(T)} \geq C_{\text{opt}}.\]

In the sequel denote by $\theta^-$ and $\theta^+$ the lower and upper bounds of the interval $\Theta : \Theta = ]\theta^-,\theta^+[$. 

\hspace{-0.6cm}The Bayes risk of $\cA$ rewrites 
\begin{eqnarray*}
\cR^H(T, \cA) &=& \bE[R_{\thetaB}(T,\cA)] = \bE\left[\sum_{a=1}^{K}(\dot{b}(\theta^*) - \dot{b}(\theta_a)) \bE_{\thetaB}[N_a(T)]\right]  \\
& = &  \sum_{a=1}^K \int_{\{\thetaB \in \Theta^K : \theta_a < \theta_a^*\}} (\dot{b}(\theta^*_a) - \dot{b}(\theta_a)) \bE_{\thetaB}[N_a(T)]dH(\thetaB)
\end{eqnarray*}
Letting $\cT_a$ be the $a$-th term in this last sum, one has 
\begin{eqnarray*}
\cT_a & = & \int_{\Theta^{K-1}}\int_{\{\theta_a \in \Theta  : \theta_a < \theta_a^*\}} (\dot{b}(\theta^*_a) - \dot{b}(\theta_a)) \bE_{\thetaB}[N_a(T)]h_a(\theta_a) d\theta_a \ dH_{-a}(\thetaa) \\
& = & \int_{\Theta^{K-1}}\int_{0}^{\theta_a^*-\theta^-}\!\!\!\!\! (\dot{b}(\theta^*_a) - \dot{b}(\theta^*_a-t)) \bE_{\thetaB_{a,t}}[N_a(T)]h_a(\theta_a^* - u)du \ dH_{-a}(\thetaa), 
\end{eqnarray*}
where $\thetaB_{a,u} := (\theta_1,\dots,\theta_{a-1},\theta_a^*-u,\theta_{a+1},\dots,\theta_K)$.

Let $\gamma \in ]0,1[$ and let $B=[b^-,b^+]$ be a compact subset of $\Theta$. For $T$ large enough, such that 
\[1/(b^- - \theta^-) < \log T < T^{\frac{1-\gamma}{2}},\]
reducing the integration domain by first letting $\thetaa \in B^{K-1}$ and then $u \in [T^{-(1-\gamma)/2}, (\log T)^{-1}]$, one has
\begin{align*}
& \cT_a  \geq    \!\!\int_{B^{K-1}}\!\!\int_{T^{-(1-\gamma)/2}}^{(\log T)^{-1}} \!\!\!(\dot{b}(\theta^*_a) - \dot{b}(\theta^*_a-u)) \bE_{\thetaB_{a,u}}[N_a(T)]h_a(\theta_a^* - u)du  dH_{-a}(\thetaa) \\
& \geq  (1-\gamma)\!\!\int_{B^{K-1}}\!\!\int_{T^{-(1-\gamma)/2}}^{(\log T)^{-1}}\!\!\frac{2h_a(\theta_a^*)\Kl(\theta_a^* -u,\theta_a^* + \zeta u)\bE_{\thetaB_{a,t}}[N_a(T)]}{u} du dH_{-a}(\thetaa).
\end{align*}
The last inequality follows from the technical lemma stated below, in which the constant $\zeta$ is defined. 

\begin{lemma}\label{lem:DL} Let $\gamma>0$. There exists $\zeta\in ]0,1[$ and $u_0>0$ such that for all $\thetaa \in B^{K-1}$ and $0\leq u \leq u_0$,
\[\forall \theta \in B, \ \ \ \ \frac{(\dot{b}(\theta)-\dot{b}(\theta-u))h_a(\theta -u)}{\Kl(\theta -u,\theta + \zeta u)} \geq (1-\gamma) \frac{2h_a(\theta)}{u}.\]
\end{lemma}

Now we need to give a lower bound on $\bE_{\thetaB_{a,u}}[N_a(T)]$, that will subsequently be integrated over $B^{K-1}\times [T^{-(1-\gamma)/2},(\log T)^{-1}]$. Lai and Robbins provide such a lower bound in \cite{LaiRobbins85bandits}, but under the assumption (not satisfied here) that, for all $\alpha \in ]0,1[$, $\cA$ has a $o(T^\alpha)$ regret on every bandit model. Moreover their lower bound is asymptotic, which makes it more complicated to integrate. Lemma~\ref{lem:LRna} below provides a non-asymptotic lower bound on  $\bE_{\thetaB_{a,u}}[N_a(T)]$, that also follows from a change of distribution argument. 

\begin{lemma} \label{lem:LRna} Let $\zeta \in ]0,1[$ and $B=[b^-,b^+] \subset \Theta$. Introducing 
\[e_{T,u}(\thetaa) = \inf \left\{\bE_{\thetaB}[T - N_a(T)] : \theta_a \in \Theta, \theta_a^* +\zeta u/2 \leq \theta_a \leq \theta_a^* + \zeta u\right\},\]
for every $\gamma \in ]0,1[$ there exists positive constants $C_1,u_1$ and $T_1$ (that depend on $B$, $\gamma$ and $\zeta$) such that if $u\leq u_1$ and $Tu^2>T_1$, $\forall \thetaa \in B^{K-1}$,
\[\bE_{\thetaB_{a,u}}[N_a(T) ] \geq \frac{(1-\gamma) \log(Tu^2)}{\Kl(\theta_a^*-u,\theta_a^*+\zeta u)}\left(1-e^{-C_1\log(Tu^2)}-\frac{2u^2 e_{T,u}(\thetaa)}{\left(Tu^2\right)^{\frac{\gamma}{2}}}\right).\]
\end{lemma}

Using Lemma~\ref{lem:LRna}, if $T$ satisfies moreover $\log(T) \geq 1/\min(u_0,u_1,\gamma/\log(T_1))$, 
\[\cT_a \geq  2(1-\gamma)^2(\cI_1(T) - \cI_2(T) - 2\cI_3(T)),\]
where
\begin{eqnarray*}
 \cI_1(T) &: = &\int_{B^{K-1}} \!\!\!h_a(\theta_a^*)\!\!\int_{T^{-(1-\gamma)/2}}^{(\log T)^{-1}}\!\!\frac{\log(Tu^2)}{u}du \ dH_{-a}(\thetaa),\\
 \cI_2(T) &:= &\int_{B^{K-1}} \!\!\!h_a(\theta_a^*)\!\!\int_{T^{-(1-\gamma)/2}}^{(\log T)^{-1}}\!\!\frac{\log(Tu^2)}{u} \frac{1}{(Tu^2)^{C_1}}du \ dH_{-a}(\thetaa),\\
 \cI_3(T)& := &\int_{B^{K-1}} \!\!\!h_a(\theta_a^*)\!\!\int_{T^{-(1-\gamma)/2}}^{(\log T)^{-1}}\!\!\frac{\log(Tu^2)}{u}t^2(Tu^2)^{-\frac{\gamma}{2}}e_{T,u}(\thetaa) du \ dH_{-a}(\thetaa).
\end{eqnarray*}

First, an explicit calculation yields 
\[\cI_1(T) =\frac{1}{4}\left(\left(1 - \frac{2\log\log(T)}{\log(T)}\right)^2 - \gamma^2\right) \log^2(T)\int_{B^{K-1}} h_a(\theta_a^*) dH_{-a}(\thetaa),\]
which shows that \[\cI_1(T) \underset{T \rightarrow \infty}{\sim} \frac{1}{4}(1-\gamma^2)\left(\int_{B^{K-1}} h_a(\theta_a^*) dH_{-a}(\thetaa)\right)\log^2(T).\]

Then, for every $\varepsilon>0$, there exists $T_2(\varepsilon)$ such that for all $T \geq T_2(\varepsilon)$, for all $u \geq T^{-(1-\gamma)/2}$, ${1}/(Tu^2)^{C_1}\leq \varepsilon$.
Hence, for $T\geq T_2(\varepsilon)$, 
\[\cI_2(T) \leq \varepsilon \cI_1(T).\]
This proves that $ \cI_2(T)= \underset{T \rightarrow \infty}{o} (\log^2(T))$.

Finally, to prove that $\cI_3(T)= \underset{T \rightarrow \infty}{o} (\log^2(T))$, we start by writing 
\[ \cI_3(T) = \int_{T^{-(1-\gamma)/2}}^{(\log T)^{-1}}\!\!\frac{\log(Tu^2)}{u}(Tu^2)^{-\frac{\gamma}{2}}u^2 \left(\!\!\int_{B^{K-1}} \!\!\!e_{T,u}(\thetaa) h_a(\theta_a^*)dH_{-a}(\thetaa)\right)   du.\]
and we provide an upper bound on the inner integral. First note that if $\thetaB$ is such that $\theta_a > \theta_a^*$, one has 
\[R_{\thetaB}(T,\cA) \geq (\dot{b}(\theta_a) - \dot{b}(\theta_a^{*}))\bE_{\thetaB}[T - N_a(T)].\] 
Using \eqref{ass:Restrict} together with this last inequality, one obtains, for every $u$, 
\begin{align*}
&C \log^2(T) \geq  %\bE\left[R_T(\cA,\thetaB)\right] \geq 
\int_{\{ \thetaB \in B^{K} : \theta_a^* + \zeta u/2<\theta_a<\theta_a^* + \zeta u\}}R_T(\cA,\thetaB)dH(\thetaB)  \\
& \hspace{1cm}\geq  \int_{B^{K-1}}\!\int_{\theta_a^* + \zeta u/2}^{\theta_a^* + \zeta u} \!\!(\dot{b}(\theta_a) - \dot{b}(\theta_a^{*}))\bE_{\thetaB}[T - N_a(T)]h_a(\theta_a)d\theta_a \ dH_{-a}(\thetaa) \\
& \hspace{1cm}\geq  \int_{B^{K-1}}e_{T,u}(\thetaa)\int_{\theta_a^* + \zeta u/2}^{\theta_a^* + \zeta u} (\dot{b}(\theta_a) - \dot{b}(\theta_a^{*}))h_a(\theta_a)d\theta_a \ dH_{-a}(\thetaa).
\end{align*}
With $B=[b^-,b^+]$, let $u_2$ be such that the compact $B' = [b^- + \zeta u_2/2, b^+ +\zeta u_2]$ is included in $\Theta$. As  $h_a$ is uniformly continuous and bounded on $B'$, there exists $u_2$ such that and for all $\theta_a^* \in B$, for all $u\leq u_2$, 
\[\inf_{[\theta_a^*+ \zeta u /2, \theta_a^*+\zeta u]} h_a(\theta) \geq {\frac{2}{3}} h_a(\theta_a^*).\]
Let $u \leq u_2$. Introducing $c_1 = \inf_{\theta \in B'}\ddot{b}(\theta)>0$, using the Lagrange formula,
\begin{eqnarray*}
C \log^2(T) &\geq & \frac{2c_1}{3}\int_{B^{K-1}}e_{T,u}(\thetaa)\int_{\theta_a^* + \zeta u/2}^{\theta_a^* + \zeta u} (\theta_a - \theta_a^{*})h_a(\theta_a^*)d\theta_a \ dH_{-a}(\thetaa) \\
& = & \frac{c_1}{4}\zeta^2u^2 \int_{B^{K-1}}e_{T,u}(\thetaa)h_a(\theta_a^*)dH_{-a}(\thetaa).
\end{eqnarray*}
Finally, if $T^{-\frac{1-\gamma}{2}} \leq u\leq u_2$, 
\[
\int_{B^{K-1}}e_{T,u}(\thetaa)h_a(\theta_a^*)dH_{-a}(\thetaa) \leq  \frac{4C}{c_1\zeta^2} \frac{\log^2(T)}{u^2} \leq \frac{4C}{c_1\zeta^2\gamma^2} \frac{(\log(Tu^2))^2}{u^2}.
\]
For $T$ satisfying $\log(T)^{-1} \leq u_2$, $[T^{-(1-\gamma)/2}, (\log T)^{-1}] \subseteq [T^{-\frac{1-\gamma}{2}}, u_2]$ and  
\begin{eqnarray*}
  \cI_3(T)& \leq &\int_{T^{-(1-\gamma)/2}}^{(\log T)^{-1}}\!\frac{\log(Tu^2)}{u}(Tu^2)^{-\frac{\gamma}{2}}u^2 \left( \frac{4C}{c_1\zeta^2\gamma^2} \frac{(\log(Tu^2))^2}{u^2} \right)  du \\
  & = & \frac{4C}{c_1\zeta^2\gamma^2}\int_{T^{-(1-\gamma)/2}}^{(\log T)^{-1}} \frac{\log(Tu^2)}{u}\left(\frac{\left(\log(Tu^2)\right)^2}{(Tu^2)^{\frac{\gamma}{2}}}\right) du.
\end{eqnarray*}
Let $\varepsilon>0$. As $x \mapsto \log^2(x)/(x^{\gamma/2})$ tends to zero when $x$ tends to infinity, and $Tu^2 \geq T^\gamma$ for $u\geq T^{-(1-\gamma)/2}$, there exists a constant $T_3(\varepsilon)$ such that 
\[\text{for}  \ T \geq T_3(\varepsilon), \text{for} \ t\geq T^{-(1-\gamma)/2}, \ \ \ \frac{\left(\log(Tu^2)\right)^2}{(Tu^2)^{\frac{\gamma}{2}}} \leq \varepsilon.\]
Hence, for $T \geq T_3(\varepsilon)$,
\begin{eqnarray*}
  \cI_3(T)  & \leq & \varepsilon \frac{4C}{c_1\zeta^2\gamma^2}\int_{T^{-(1-\gamma)/2}}^{(\log T)^{-1}} \frac{\log(Tu^2)}{u} du\\
  &= &\varepsilon \frac{C}{c_1\zeta^2\gamma^2}\left(\left(1 - \frac{2\log\log(T)}{\log(T)}\right)^2 - \gamma^2\right) \log^2(T),
\end{eqnarray*}
which proves that $\cI_3(T) = o\left(\log^2(T)\right).$

Putting everything together, we proved that, for every algorithm $\cA$, for every $\gamma>0$, for every compact $B \subset\Theta$, 
\[\liminf_{T \rightarrow \infty} \frac{\cR_T(\cA,H)}{\log^2(T)} \geq (1-\gamma)^2(1-\gamma^2)\frac{1}{2}\sum_{a=1}^K\int_{B^{K-1}} h_a(\theta_a^*) dH_{-a}(\thetaa).\]
Taking the supremum over all compact set $B$ yields, for every $\gamma>0$, 
\[\liminf_{T \rightarrow \infty} \frac{\cR_T(\cA,H)}{\log^2(T)} \geq  (1-\gamma)^2(1-\gamma^2)\frac{1}{2}\sum_{a=1}^K\int_{\Theta^{K-1}} h_a(\theta_a^*) dH_{-a}(\thetaa),\]
provided the integral in the right-hand side is finite. Letting $\gamma$ go to zero concludes the proof.

\subsection{Proof of Lemma~\ref{lem:DL}}

Let $\zeta \in ]0,1[$ be fixed . As $B = [b^-,b^+]$ is strictly included in $\Theta$, there exists $u_1$ such that $\cC:=[b^- - u_1, b^+ + \zeta u_1]$ in included in $\Theta$. 
For $(\theta,u) \in B \times [0,u_1]$ we define  
\[f(\theta,u) = \frac{(1+\zeta)^2u}{2}\frac{(\dot{b}(\theta) - \dot{b}(\theta-u))h_a(\theta -u)}{K(\theta - u, \theta + \zeta u)h_a(\theta)}.\]
$f$ is continuous on $B \times [0,t_1]$ and it can be checked that 
\[\lim_{(\theta,u)\rightarrow (\theta_0,0)} f(\theta,u) = 1.\]
As $f$ is uniformly continuous, there exists  $u_0 \leq u_1$, such that for all $u\leq u_0$, for all $\theta \in B$, 
\[|f(\theta,u) - 1 | \leq \frac{\gamma}{2},\]
which rewrites 
\[
\left|\frac{(\dot{b}(\theta) - \dot{b}(\theta-u))h_a(\theta -u)}{K(\theta - u, \theta + \zeta u)} - \frac{2h_a(\theta)}{(1+\zeta)^2u}\right| \leq \frac{\gamma}{2} \frac{2h_a(\theta)}{(1+\zeta)^2u}
\]
hence, for $u\leq u_0$, one has 
\[
\frac{(\dot{b}(\theta) - \dot{b}(\theta-u))h_a(\theta -u)}{K(\theta - u, \theta + \zeta u)}\geq \frac{1-\frac{\gamma}{2}}{(1+\zeta)^2}\frac{2h_a(\theta)}{u}.
\]
Applying this to $\zeta$ such that $1 + \zeta = \sqrt{\frac{1-\frac{\gamma}{2}}{1-\gamma}}$ concludes the proof.

\subsection{Proof of Lemma~\ref{lem:LRna}}

Let $\zeta \in ]0,1[$ be fixed and define $u_1$ and $\cC=[b^- - u_1, b^+ + \zeta u_1] \subset \Theta$ as in the proof of Lemma~\ref{lem:DL}. Let $u\leq u_1$ and fix $\thetaa \in B^{K-1}$.  First, using Markov inequality, 
\[\bE_{\thetaB_{a,u}}[N_a(T) ] \geq \frac{(1-\gamma) \log(Tu^2)}{K(\theta_a^*-u,\theta_a^*+\zeta u)}\bP_{\thetaB_{a,u}}\left(N_a(T) \geq \frac{(1-\gamma) \log(Tu^2)}{K(\theta_a^*-u,\theta_a^*+\zeta u)}\right).\]
Thus it is sufficient to prove that there exists $C_1>0$ such that
\begin{equation}\bP_{\thetaB_{a,u}}\left(N_a(T) \leq \frac{(1-\gamma) \log(Tu^2)}{K(\theta_a^*-u,\theta_a^*+\zeta u)}\right) \leq e^{-C_1\log(Tu^2)}+ \frac{2u^2e_{T,u}(\thetaa)}{\left(Tu^2\right)^{\frac{\gamma}{2}}}.\label{step:ToProve}\end{equation}

As $u\leq u_1$, the set $\left\{ \theta_a : \theta_a^* +\zeta u/2 \leq \theta_a \leq \theta_a^* + \zeta u\right\}$
is a compact set included in $\cC$, therefore there exists $\bm{\lambda} \in B^{K-1}\times \cC$ that attains the infimum in the definition of $e_{T,u}(\thetaa)$:
\[e_{T,u}(\thetaa) = \bE_{\bm{\lambda}}[T - N_a(T)],\]
with $\bm{\lambda}_{-a} = \thetaa$ and $\lambda_a = \theta_a^* + \varepsilon u$, for some $\varepsilon \in [\zeta/2,\zeta]$. Using Markov inequality, 
\begin{eqnarray*}
 \bP_{\bm\lambda}\left(N_a(T) \leq \frac{(1-\gamma) \log(Tu^2)}{K(\theta_a^*-u,\theta_a^*+\zeta u)}\right) &\leq& \frac{\bE_{\bm{\lambda}}[T - N_a(T)]}{T - \frac{(1-\gamma) \log(Tu^2)}{K(\theta_a^*-u,\theta_a^*+\zeta u)}}=\frac{e_{T,u}(\thetaa)}{T\left(1-\frac{(1-\gamma) \log(Tu^2)}{K(\theta_a^*-u,\theta_a^*+\zeta u)T}\right)}.
\end{eqnarray*}
Introducing $c_1 = \inf_{\theta \in \cC} \ddot{b}(\theta)$, using \eqref{LagrangeK} in Proposition~\ref{prop:Pinsker}, for $u\leq u_1$,   
\[\frac{(1-\gamma) \log(Tu^2)}{K(\theta_a^*-u,\theta_a^*+\zeta u)T} \leq \frac{2(1-\gamma) \log(Tu^2)}{c_1(1+\zeta)^2(Tu^2)} \leq \frac{1}{2},\]
where the last inequality holds to $Tu^2$ large enough. Thus there exists $T_1>0$ such that for $u\leq u_1$ and  $Tu^2 \geq T_1$,  
\begin{equation}\bP_{\bm\lambda}\left(N_a(T) \leq \frac{(1-\gamma) \log(Tu^2)}{K(\theta_a^*-u,\theta_a^*+\zeta u)}\right) \leq \frac{2e_{T,u}(\thetaa)}{T}.\label{step:Lambda}\end{equation}

Introducing the log likelihood ratio $L_n = \sum_{s=1}^{n} \log \frac{f_{\theta^*_a-t}(Y_{a,s})}{f_{\lambda_a}(Y_{a,s})}$, where $Y_{a,s}$ are i.i.d. samples of the distribution of arm $a$, one can write
\begin{align}
&\bP_{\thetaB_{a,u}}\left(N_a(T) \leq \frac{(1-\gamma) \log(Tu^2)}{K(\theta_a^*-u,\theta_a^*+\zeta u)}\right)  \nonumber \\
& \leq   \bP_{\thetaB_{a,u}}\left(N_a(T) \leq \frac{(1-\gamma) \log(Tu^2)}{K(\theta_a^*-u,\theta_a^*+\zeta u)} , L_{N_a(T)} \leq \left(1-\frac{\gamma}{2}\right)\log (Tu^2)\right)\label{step:Term1} \\
& \hspace{0.5cm} + \bP_{\thetaB_{a,u}}\left(\max_{n \leq \frac{(1-\gamma) \log(Tu^2)}{K(\theta_a^*-u,\theta_a^*+\zeta u)}} L_n \geq \left(1-\frac{\gamma}{2}\right)\log (Tu^2)\right)\label{step:Term2}
\end{align}

An upper bound on Term~\eqref{step:Term1} follows from a change of distribution argument. Let $\cE$ be the event 
\[\cE := \left\{N_a(T) \leq \frac{(1-\gamma) \log(Tu^2)}{K(\theta_a^*-u,\theta_a^*+\zeta u)} , L_{N_a(T)} \leq \left(1-\frac{\gamma}{2}\right)\log (Tu^2) \right\}\]
As $\cE \in \cF_{N_a(T)}$, one has 
\[
\bP_{\bm\lambda}(\cE) = \bE_{\thetaB_{a,u}}\left[\ind_{\cE} \exp\left(-L_{N_a(T)}\right)\right] \geq \exp\left(-\left(1-\frac{\gamma}{2}\right)\log (Tu^2) \right)  \bP_{\thetaB_{a,u}}(\cE).
\]
Thus, using moreover \eqref{step:Lambda}, 
\begin{eqnarray*}
\eqref{step:Term1} &\leq& (Tu^2)^{1-\frac{\gamma}{2}} \bP_{\bm\lambda}(\cE) \leq (Tu^2)^{1-\frac{\gamma}{2}} \bP_{\bm\lambda}\left(N_a(T) \leq \frac{(1-\gamma) \log(Tu^2)}{K(\theta_a^*-u,\theta_a^*+\zeta u)}\right) \\
&\leq& 2 u^2 \left(Tu^2\right)^{-\frac{\gamma}{2}} e_{T,u}(\thetaa). 
\end{eqnarray*}

An upper bound  of Term~\eqref{step:Term2} follows from a concentration inequality specific to exponential families, stated as Lemma~\ref{lem:ConcExpo}, whose proof is provided below for the sake of completeness. 

\begin{lemma}\label{lem:ConcExpo} Let the $(Y_i)$ be i.i.d with distribution $\nu_{\theta}$ and mean $\mu=\dot{b}(\theta)$. 
\[\bP\left(\max_{ n \leq N} \sum_{s=1}^n (\mu - Y_i) \geq x \right) \leq \exp\left(- N d\left(\mu - \frac{x}{N},\mu\right)\right)\] 
\end{lemma}

Introducing the notation $\overline{\theta}_a = \theta_a^* -u$ and  
\[K_T = \frac{(1-\gamma) \log(Tu^2)}{K(\theta_a^*-u,\theta_a^*+\zeta u)} = \frac{(1-\gamma) \log(Tu^2)}{K(\overline{\theta}_a,\theta_a^*+\zeta u)},\]
the log likelihood ratio can be made explicit, and satisfies, for $n \leq K_T$, 
\begin{eqnarray*}
 L_n & = & \sum_{s=1}^n (\overline{\theta}_a -\lambda_a)Y_{a,s} - b(\overline{\theta}_a)+b(\lambda_a) \\
 &= &(\overline{\theta}_a -\lambda_a)\sum_{s=1}^n (Y_{a,s} - \dot{b}(\overline{\theta}_a)) + n K(\overline{\theta}_a,\lambda_a). \\
 & \leq & (\lambda_a-\overline{\theta}_a)\sum_{s=1}^n (\dot{b}(\overline{\theta}_a)-Y_{a,s}) + (1-\gamma)\log(Tu^2).
\end{eqnarray*}
Term \eqref{step:Term2} is upper bounded by 
\begin{align*}
 &\bP_{\thetaB_{a,u}}\left(\!\max_{n \leq K_T}\!\left[\!(\lambda_a \! -\overline{\theta}_a)\! \sum_{s=1}^n (\dot{b}(\overline{\theta}_a)\!-\!Y_{a,s})\! + \! (1-\gamma)\log(Tu^2)\right]\!\!\geq \left(\!1-\!\frac{\gamma}{2}\right)\log (Tu^2)\!\right) \\
 & \leq  \bP_{\thetaB_{a,u}}\left(\max_{n \leq K_T} \sum_{s=1}^n (\dot{b}(\overline{\theta}_a)-Y_{a,s} )\geq \frac{\gamma}{2}\frac{\log (Tu^2)}{\lambda_a-\overline{\theta}_a}\right). 
\end{align*}
Under $\thetaB_{a,u}$, the sequence $Y_{a,s}$ is i.i.d with distribution $\nu_{\overline{\theta}_a}$. Therefore, using Lemma~\ref{lem:ConcExpo} one obtains, with the notation $\overline{\mu}_a = \dot{b}(\overline{\theta}_a)$,
\begin{eqnarray*}
  \eqref{step:Term2} & \leq &\exp\left(- K_T d\left(\overline{\mu}_a - \frac{\gamma K(\theta^*_a - t,\theta_a^* + \zeta u)}{2(1-\gamma)(\varepsilon +1) u},\overline{\mu}_a\right)\right).
\end{eqnarray*}
Letting $c_1 = \inf_{\theta \in \cC} \ddot{b}(\theta)$ and $c_2 = \inf_{\theta \in \cC} \ddot{b}(\theta)$, from \eqref{LagrangeK} in Proposition~\ref{prop:Pinsker},
\[\frac{\gamma K(\theta^*_a - u,\theta_a^* + \zeta u)}{2(1-\gamma)(\varepsilon +1) u} \in \left[ \frac{\gamma}{2(1-\gamma)}\frac{c_1}{2}(\zeta + 1)^2 u^2 ; \frac{\gamma}{2(1-\gamma)}\frac{c_2}{2}(\zeta + 1)^2 u^2 \right].\]
Thus, for $u$ small enough, $\overline{\mu}_a$ and $\overline{\mu}_a -\frac{\gamma K(\theta^*_a - u,\theta_a^* + \zeta u)}{2(1-\gamma)(\varepsilon +1) u}$ belong to a compact $\cC'$ satisfying $\cC \subseteq \cC' \subseteq \Theta$. Letting $c_2' = \sup_{\theta \in \cC'} \ddot{b}(\theta)$, using \eqref{LagrangeD}, 
\begin{eqnarray*}
 \eqref{step:Term2}  &\leq& \exp\left(- \frac{K_T}{2c'_2} \left(\frac{\gamma K(\theta^*_a - u,\theta_a^* + \zeta u)}{2(1-\gamma)(\varepsilon +1) u}\right)^2\right) \\
  & = & \exp\left(- \log(Tu^2) \frac{\gamma^2}{8(1-\gamma)c'_2}\frac{K(\theta_a^* -u,\theta_a^* + \zeta u)}{(1+\varepsilon)^2u^2}\right) \\
  & \leq & \exp\left(- \log(Tu^2) \frac{\gamma^2c_1}{8(1-\gamma)c'_2}\frac{c_1(1+\zeta)^2}{(1+\varepsilon)^2}\right).
\end{eqnarray*}
Letting $C_1 = \frac{\gamma^2c_1}{8(1-\gamma)c_2'}\frac{c_1(1+\zeta)^2}{(1+\varepsilon)^2}$, from the upper bounds obtained on \eqref{step:Term1} and \eqref{step:Term2}, it follows that 
\[\bP_{\thetaB_{a,u}}\left(N_a(T) \leq \frac{(1-\gamma) \log(Tu^2)}{K(\theta_a^*-u,\theta_a^*+\zeta u)}\right) \leq 2 u^2\left(Tu^2\right)^{-\frac{\gamma}{2}} e_{T,u}(\thetaa) + e^{-C_1\log(Tu^2)},\]
provided that $u\leq u_1$ and $Tu^2 \geq T_1$, which concludes the proof. \qed 

\paragraph{Proof of Lemma~\ref{lem:ConcExpo}} The proof follows from the Chernoff technique and a maximal inequality. 

Let $S_n = \sum_{s=1}^n (\mu - Y_i)$. For every $\lambda >0$, 
\begin{equation}\bP\left(\max_{n \leq N} S_n \geq x \right) = \bP\left(\max_{n \leq N} e^{\lambda S_n} \geq e^{\lambda x} \right) \leq e^{-\lambda x} \bE\left[e^{\lambda S_N}\right],\label{equ:Doob}\end{equation}
where the last inequality is a consequence of Doob's maximal inequality applied to $M_n = e^{\lambda S_n},$ which is a sub-martingale with respect to the filtration generated by the $(Y_i)$. Indeed, using the convexity of the mapping $x \mapsto e^{\lambda x}$, 
\begin{eqnarray*}
\bE\left[ M_n - M_{n-1} | \cF_{n-1} \right]  & = &  e^{\lambda S_n} \bE\left[e^{\lambda (S_n - S_{n-1})} -1 | \cF_{n-1}\right] \\ 
& \geq & e^{\lambda S_n} \lambda \bE\left[S_n - S_{n-1}| \cF_{n-1}\right] =0. 
\end{eqnarray*}
Using the independence of the $Y_i$ and $\bE[e^{\lambda Y_i}] = \exp(b(\theta + \lambda) - b(\theta)$ for any $\lambda \in \R$, it can be show that 
\[e^{-\lambda x} \bE\left[e^{\lambda S_N}\right] = \exp\left(-N \left[\lambda\left(\frac{x}{N} - \dot{b}(\theta)\right) + b(\theta) - b(\theta - \lambda)\right]\right).\]
The exponent is minimized of $\lambda^*$ satisfying 
$\dot{b}(\theta - \lambda^*) = \dot{b}(\theta) - {x}/{N}$
and 
\begin{eqnarray*}
e^{-\lambda^* x} \bE\left[e^{\lambda^* S_N}\right] & = & \exp\left(-N \left[\dot{b}(\theta - \lambda^*) (-\lambda^*) - \dot{b}(\theta - \lambda^*)+ b(\theta)\right]\right) \\ 
& = & \exp(-N K(\theta - \lambda^*, \theta)) = \exp\left( - N d\left(\mu - \frac{x}{N},\mu \right)\right).
\end{eqnarray*}
The conclusion follows by plugging $\lambda^*$ in \eqref{equ:Doob}.

\subsection{The lower bound for Bernoulli bandits\label{subsec:BorneBernoulli}}

As pointed out by \cite{Lai87}, in the particular case in which $h_a(\theta)=q(\theta)$ for all $a=1,\dots,K$, using the fact that the distribution of $\max_{a \in \cS} \ \theta_a$ has density $kq(\theta)Q^{k-1}(\theta)$ where $Q$ is the c.d.f. of the distribution with density $q$ and $k=|\cS|$, the constant in the lower bound can be expressed
\begin{equation}\frac{1}{2}\sum_{a=1}^K\int_{\Theta^{K-1}} h_a(\theta_a^*) dH_{-a}(\thetaa) = \frac{K(K-1)}{2}\int_{\Theta} q^2(\theta) Q^{K-2}(\theta)d\theta\label{version:Homogene}.\end{equation}
Now consider a Bernoulli bandit model with $K$ arms, with a uniform prior distribution on the mean of each arm. The set of Bernoulli distribution of means $\mu \in [0,1]$ form an exponential family when each distribution is parametrized by the natural parameter $\theta=\log(\mu/(1-\mu))$. This exponential family is characterized by  
\[\Theta = \R, \ \ b(\theta) = \log(1+e^\theta),\]
and the reference measure is the Lebesgue measure.  As each mean $\mu_a$ is drawn from a uniform distribution on $[0,1]$, the associated natural parameter $\theta_a$ is drawn from a distribution on $\R$ having respectively density and c.d.f. 
\[q(\theta)=\frac{e^\theta}{(1+e^\theta)^2} \ \ \ \text{and} \ \ \ \ Q(\theta) = \frac{e^\theta}{1+e^\theta}.\]
Using the formula \eqref{version:Homogene}, the constant of the lower bound is 
\begin{eqnarray*}
 \frac{K(K-1)}{2}\int_{-\infty}^{+\infty} \frac{e^{K\theta}}{(1+e^\theta)^{K+2}} d\theta & =&  \frac{K(K-1)}{2}\int_0^\infty \frac{x^{K-1}}{(1+x)^{K+2}}dx \\
 &=& \frac{K(K-1)}{2} \frac{1}{K(K+1)},
\end{eqnarray*}
where the integral is computed using by inducting, using a by part integration. Finally, the asymptotic rate of the Bayes risk for a Bernoulli bandit model with $K$ arms and a uniform prior on their means is  
\[\frac{1}{2}\frac{K-1}{K+1}\log^2(T).\]

\end{document}